\documentclass[10pt,twocolumn,letterpaper]{article}
 
\usepackage{include/template/cvpr}              

\usepackage{epsfig}
\usepackage{graphicx}
\usepackage{amsmath}
\usepackage{amssymb}
\usepackage{booktabs}
\usepackage{caption}
\usepackage{subcaption}
\usepackage{xcolor,colortbl}
\usepackage[accsupp]{axessibility}

\usepackage[pagebackref,breaklinks,colorlinks]{hyperref}

\usepackage[capitalize]{cleveref}
\crefname{section}{Sec.}{Secs.}
\Crefname{section}{Section}{Sections}
\Crefname{table}{Table}{Tables}
\crefname{table}{Tab.}{Tabs.}
\def\ours{{BrAD}}
\def\oursspace{{BrAD }}
\def\oursfull{Bridge Across Domains }

\def\ourstask{UDG}
\def\ourstaskspace{UDG }
\def\ourstaskfull{Unsupervised Domain Generalization }

\newcommand\splus{\vcenter{\hbox{\scalebox{1}{+}}}}
\newcommand\sminus{\vcenter{\hbox{\scalebox{1}{-}}}}

\definecolor{darkgreen}{rgb}{0.0,0.6,0.0}

\newcommand\blue[1]{\textcolor{blue}{#1}}
\newcommand\red[1]{\textcolor{red}{#1}}
\definecolor{cyan}{rgb}{0.45,0.87,0.95}
\definecolor{orange}{rgb}{0.95,0.8,0.6}
\definecolor{britishracinggreen}{rgb}{0.0, 0.26, 0.15}
\definecolor{cadmiumgreen}{rgb}{0.0, 0.42, 0.24}

\newcommand\secvspace{\vspace{-0.2cm}}

\newcommand\figvspace{\vspace{-0.5cm}}
\newcommand\tabvspace{\vspace{-0.6cm}}

\newcommand\democap[1]{Random query example from our demo. For each query image (from PACS) we show the top $5$ image matches among the entire set of images in each of the $4$ PACS domains: Photo, Art/Painting, Cartoon, and Sketch. The matching is obtained using our self-supervised \oursspace model trained using DomainNet data. The correct class is \red{#1}. The text under each image is the ground truth class of that image in the PACS dataset.}

\newcommand\blfootnote[1]{%
  \begingroup
  \renewcommand\thefootnote{}\footnote{#1}%
  \addtocounter{footnote}{-1}%
  \endgroup
}

\def\cvprPaperID{3951} 
\def\confName{CVPR}
\def\confYear{2022}

\begin{document}

\title{\ourstaskfull by Learning a \oursfull}

\author{
    Sivan Harary*$^{1}$,
    Eli Schwartz*$^{1,2}$,
    Assaf Arbelle*$^{1}$,\\
    Peter Staar$^{1}$,
    Shady Abu-Hussein$^{1,2}$,
    Elad Amrani$^{1,3}$,
    Roei Herzig$^{1,2}$,\\
    Amit Alfassy$^{1,3}$,
    Raja Giryes$^{2}$,
    Hilde Kuehne$^{6,7}$,
    Dina Katabi$^{5}$,\\
    Kate Saenko$^{4,7}$,
    Rogerio Feris$^{7}$,
    Leonid Karlinsky\thanks{Equal contribution}\hspace{-1pt}*$^{1}$\\
    \\
    \tt\small 
    $^{1}$IBM Research, 
    $^{2}$Tel-Aviv University,  
    $^{3}$Technion,
    $^{4}$Boston University, \\
    \tt\small 
    $^{5}$MIT,
    $^{6}$Goethe University,
    $^{7}$MIT-IBM Watson AI Lab
}



\maketitle

\begin{abstract}
\vspace{-0.2cm}
   The ability to generalize learned representations across significantly different visual domains, such as between real photos, clipart, paintings, and sketches, is a fundamental capacity of the human visual system. In this paper, different from most cross-domain works that utilize some (or full) source domain supervision, we approach a relatively new and very practical \ourstaskfull(\ourstask) setup of having \textbf{no training supervision} in neither source nor target domains. Our approach is based on self-supervised learning of a \oursfull(\ours) - an auxiliary \textbf{bridge} domain accompanied by a set of semantics preserving visual (image-to-image) mappings to \oursspace from each of the training domains. The \oursspace and mappings to it are learned jointly (end-to-end) with a contrastive self-supervised representation model that semantically aligns each of the domains to its \ours-projection, and hence implicitly drives all the domains (seen or unseen) to semantically align to each other. In this work, we show how using an edge-regularized \oursspace our approach achieves significant gains across multiple benchmarks and a range of tasks, including UDG, Few-shot UDA, and unsupervised generalization across multi-domain datasets (including generalization to unseen domains and classes).
\end{abstract}


\vspace{-0.3cm}
\secvspace
\section{Introduction}
\secvspace

\begin{figure}[t]
\vspace{-0.4cm}
\begin{center}
   \includegraphics[width=0.9\columnwidth]{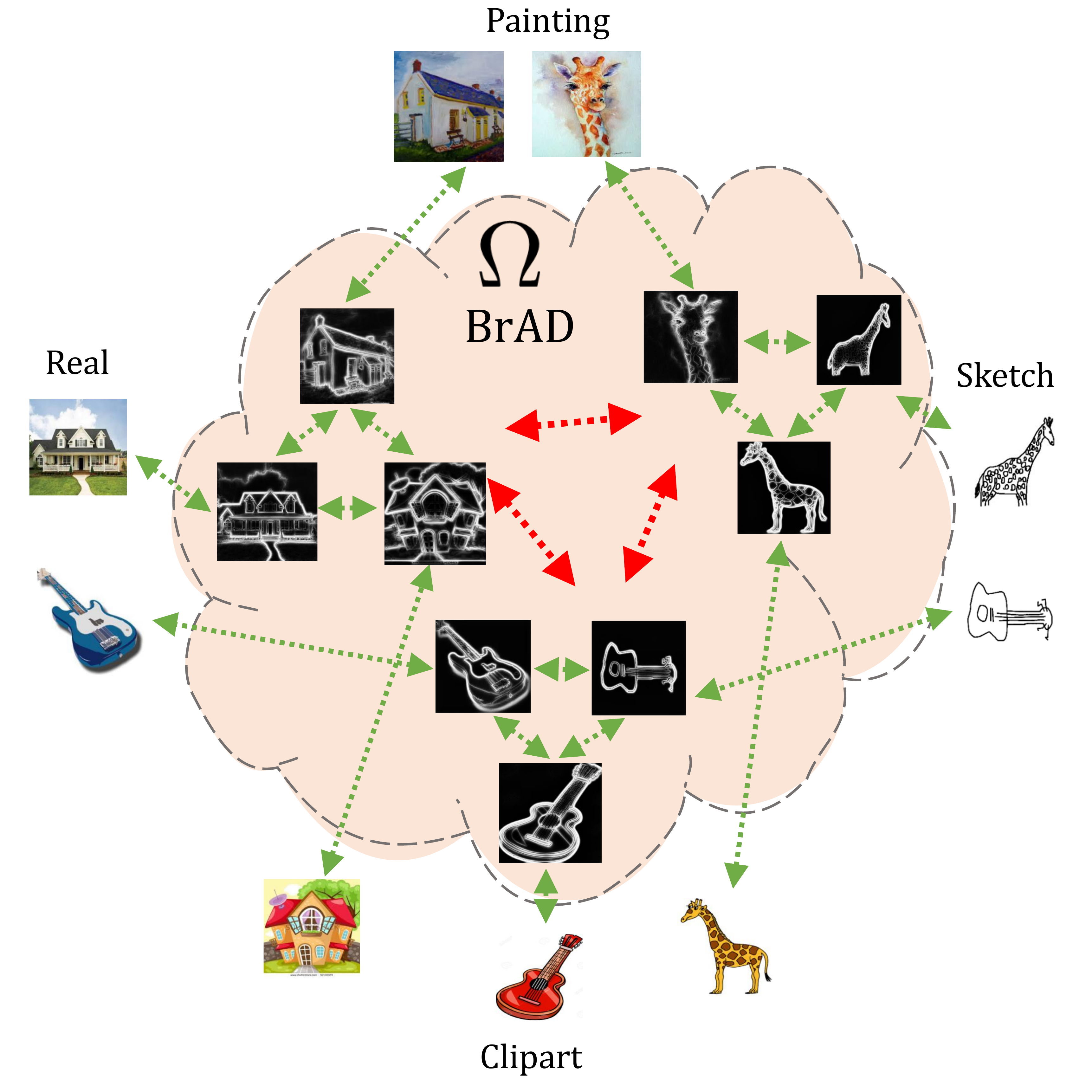}
\end{center}
\vspace{-20pt}
   \caption{Originally, same domain instances are closer to each other than to instances of the same class in other domains. Naive application of popular self-supervised learning techniques tends to separate domains before classes (Section \ref{sec:results}). Our approach is to learn a \oursspace - an auxiliary bridge domain (e.g. of edge-like images) that helps aligning instances of the same class across domains. Arrows indicate forces in feature space applied by our training losses (Section \ref{sec:method}). \textbf{\textcolor{cadmiumgreen}{green}} = attraction; \textbf{\textcolor{red}{red}} = repulsion.
}
\label{fig:intro}
\figvspace
\end{figure}

%
%
When observing some technical manual with schematic drawings of equipment for the first time, people do not need any supervision to associate these schematic drawings to real complex objects. This demonstrates the importance and efficiency of one of the basic human capacities - ability to learn \textit{with little to no supervision} across multiple visual domains, as well as to effectively \textit{generalize to new domains} and even new object classes without additional supervision.

\blfootnote{
\hspace{-14pt}\textbf{corresponding authors:} sivangl$\|$elisch$\|$assaf.arbelle$\|$leonidka@il.ibm.com
}
Recent literature, extensively discussed in Section~\ref{sec:related}, is rich in works on learning to semantically generalize across domains without supervision in the target domain(s). The target domains can be observed as a collection of unlabeled images in Unsupervised Domain Adaptation (UDA), or even completely unseen during training in Domain Generalization (DG). For both UDA and DG, success in generalization would mean that the desired downstream tasks (classification, detection, etc.) would successfully transfer to a new seen or unseen domain. However, in most UDA and DG works an abundant source domain supervision for the intended downstream task is assumed. But do we always have that in real-life use-cases? In many situations, such as in the technical manuals example above, we need our systems to generalize not only to new domains, but also to completely new kinds of objects for which we might have very little data (images and/or labels) - not sufficient to train the standard UDA or DG methods. Few recent Unsupervised DG (UDG) and Few-Shot UDA (FUDA) works, realized the importance of this issue, and restrict the source domain(s) supervision to few or even zero labeled examples. In this paper we target the least restrictive (in terms of labeling requirements) \ourstaskspace setting, where we do not require any source domain supervision at training, and which also implicitly supports generalization to completely new unseen visual domains with new unseen classes.

We all draw an edge-like image when asked to draw something quickly. Object edges seems to be our shared universal visual representation for all the domains we observe. This drove the basic intuition behind our \oursspace approach: teach the machine to represent all the visual domains (in feature space) in the same way as it represents this seemingly universally shared visual domain of edge-like images. Our approach (Figure~\ref{fig:intro}) is based on our proposed concept of a learnable \oursfull(\ours) - an auxiliary visual `bridge' domain to which it is relatively easy to visually map (in an image-to-image sense) all the domains of interest. The \oursspace is used only during contrastive self-supervised training of our model for semantically aligning the representations (features) of each of the training domains to the ones for the shared \ours. By transitivity of this semantic alignment in feature space, the learned model representations for all the domains are all \ours-aligned and hence implicitly aligned to each other. This makes the learned mappings to the \oursspace unnecessary for inference, making the trained model generalize-able even to new unseen domains (for which we do not have \oursspace mappings) at test time. Moreover, not depending on \ours-mapping for inference allows exploiting additional non-\ours-specific features (e.g. color) which are also learned by our representation model encoder (\cref{sec:method}). We show that even a simple heuristic implementation of the \oursspace idea, mapping images to their edge maps, already gives a nice improvement over strong self-supervised baselines not utilizing \ours. We further show that with learnable \ours, our method demonstrates good gains across various datasets and tasks: UDG, FUDA, and a proposed task of generalization to different domains and classes.

To summarize our contributions are as follows: \textbf{\blue{(i)}} we propose a new concept of a learnable \oursspace- an auxiliary visual `bridge' domain which is relatively easily mapable from domains of interest, seen or unseen, and can drive the learned representation features to be semantically aligned (generalize) across domains; \textbf{\blue{(ii)}} we show how using the concept of \oursspace combined with self-supervised contrastive learning and some additional ideas allows to train an effective (and efficient) model for different kinds of source-labels-limited cross-domain tasks, including UDG, FUDA, and even generalization across multi-domain benchmarks without supervision; \textbf{\blue{(iii)}} we demonstrate significant gains of up to $14\%$ over UDG and up to $13.3\%$ over FUDA respective state-of-the-art (SOTA) on several benchmarks, as well as showing significant advantages in transferring the learned representations without any additional fine-tuning to new unseen domains and object categories.

\begin{figure*}[t]
\vspace{-0.4cm}
\begin{center}
  \includegraphics[width=0.9\linewidth]{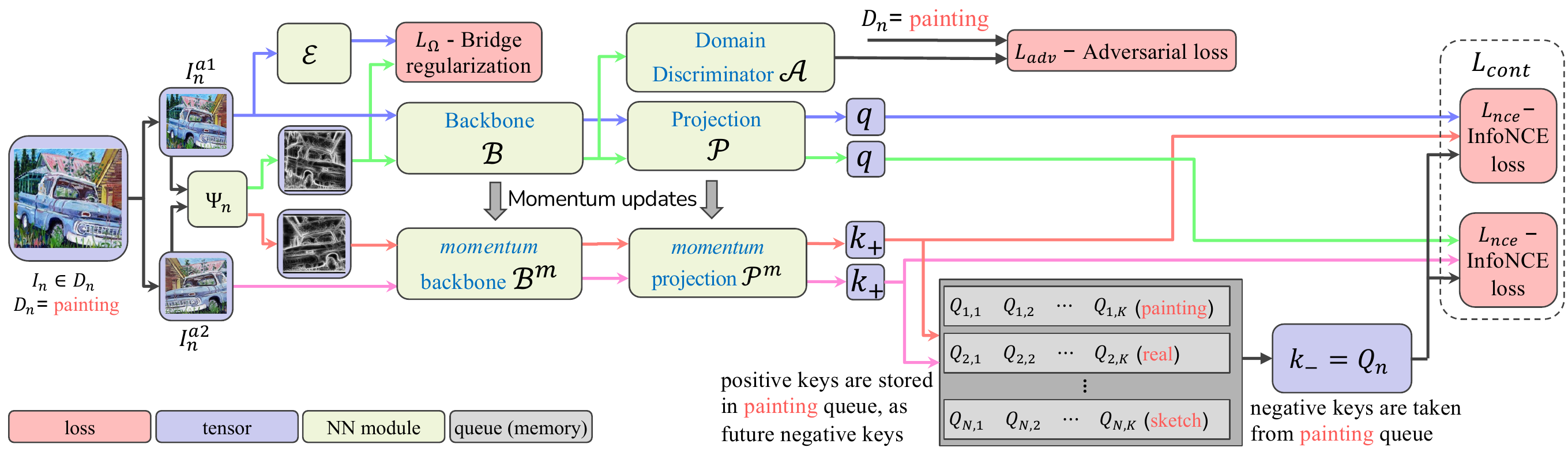}
\end{center}
\vspace{-15pt}
   \caption{\textit{\oursspace training architecture.} 
   $I_n^{a1},I_n^{a2}$ are random augmentations of an image $I_n \in D_n$. $\Psi_n(I_n^{a1}),\Psi_n(I_n^{a2})$ are their respective mappings to the bridge domain, $\Omega$, using a (learned) domain specific image-to-image mapping $\Psi_n$. 
   The color of the arrows indicates the image that flows through the model: $I_n^{a1}$ (blue), $I_n^{a2}$ (pink), $\Psi_n(I_n^{a1})$ (green) and $\Psi_n(I_n^{a2})$ (red).
   The negative keys $k_-$ of the contrastive losses $L_{nce}$ come from a domain specific queue. We apply the $L_\Omega$ regularization, distilling from an edge-mapping $\mathcal{E}$ (can be even a simple Canny, see \cref{sec:ablations}), forcing the bridge domain images to be similar to edge maps, which are (intuitively) less sensitive to domain shift. Finally, a domain discriminator $\mathcal{A}$ and an adversarial loss $\mathcal{L}_{adv}$ improve the domain invariance of the $\Omega$ projected images representations. 
}
\label{fig:arch}
\figvspace
\end{figure*}
\secvspace
\section{Related Work}\label{sec:related}
\secvspace


\noindent\textbf{Unsupervised Domain Adaptation (UDA).} UDA~\cite{gopalan2011domain} refers to transferring knowledge from a \textit{labeled} source domain to an \textit{unlabeled} target domain. 
Most UDA methods employ feature distribution alignment using: maximum discrepancy of domain distributions~\cite{Saito_2018_CVPR,zhang2019bridging,long2015learning,long2016unsupervised,sun2017correlation,zhuo2017deep,kang2020enhanced}, adversarial domain classifier~\cite{ganin2016domain,hoffman2018cycada,Long18,ganin2015unsupervised,Tzeng_2017_CVPR,xie2018learning,shen2018wasserstein}, and entropy optimization~\cite{Saito19,Long18,kim2021cds,saito2020universal}. GAN based image translation is used in~\cite{Bousmalis_2017_CVPR,hoffman2018cycada,Murez_2018_CVPR,Sankaranarayanan_2018_CVPR,Shrivastava_2017_CVPR,Lee_2021_CVPR}. In \cite{liang2020we,wang2021cross,yang2020unsupervised} source domain data is replaced with a model pre-trained on the source. \cite{lopez2020desc} uses low-level edge features to enforce consistency in UDA for monocular depth estimation. In Few-shot UDA (FUDA)~\cite{kim2021cds,Yue2021}, only a few examples per class are labeled in the source domain, while the rest are unlabeled. In our work we take another step further and preform domain adaptation without seeing any labeled examples during training.

\noindent\textbf{Domain Generalization (DG).}
DG addresses transferring knowledge to domains unseen during training. Most DG works perform supervised training on (a set of) labeled source domains. Methods for supervised DG include: distribution matching across domains \cite{li2018domain, li2018deep}, adaptive weighting of different source domain losses \cite{sagawa2019distributionally}, enforcing low rank of the latent representations \cite{li2020domain} , and random source domains mixing \cite{wang2020heterogeneous}. Unsupervised DG (UDG) is a new task of training with unlabeled source domains introduced by~\cite{zhang2021domain}. The unlabeled source domains images are used for self-supervised pretraining followed by fitting a classifier to the learned representations via labeling a small portion of the source images. We show strong advantages of our method for the UDG setting, even when the unsupervised pretraining features are used directly via kNN without any further classifier fitting.

\noindent\textbf{Self-Supervised Learning (SSL).}
SSL refers to learning strong semantic feature representations from unlabeled data. Historically, many pre-text tasks were proposed for SSL
~\cite{dosovitskiy2014discriminative,noroozi2016unsupervised,DBLP:conf/iclr/GidarisSK18,zhang2016colorful,pathak2016context,koohpayegani2021mean}. Recently, contrastive learning ~\cite{dosovitskiy2014discriminative,wu2018unsupervised,DBLP:journals/corr/abs-2002-05709,chen2020big,he2020momentum,chen2020improved,caron2020unsupervised,grill2020bootstrap, chen2021exploring, caron2021emerging,kotar2021contrasting,tejankar2021isd,wang2021solving, zbontar2021barlow} has shown great promise and SOTA results. 
Contrastive methods usually optimize instance discrimination by making two augmentations of the same image closer in feature space than their distance to a set of negative anchors. 
In our experiments we show an advantage of our method over a direct application of popular SSL methods to multi-domain data, likely due to SSL methods tendency to separate domains before separating classes, avoided via the proposed use of a cross-domain bridge. Concurrently to our work, \cite{chenwei2021} successfully used HOG features for SSL in videos. Our approach can easily allow adoption of HOG or other hand-crafted edge-based features as an alternative regularization for the learned bridge domain, thus offering a general tool for further investigation of edges utility for contrastive SSL in future work. 



\noindent\textbf{Self-Supervised Learning for UDA and DG.}
Some UDA and DG methods employ SSL losses in their pipeline. 
SSL task of solving jigsaw puzzle was leveraged in~\cite{noroozi2016unsupervised} to aid UDA and DG. ~\cite{sun2019unsupervised} combine supervised learning on the source domain jointly with SSL on both domains. Recently~\cite{kim2021cds,Yue2021} proposed a cross-domain SSL approach for Few-shot UDA (FUDA), and \cite{zhang2021domain} proposed an SSL method for UDG. As mentioned, we show strong advantages of our method for both FUDA and UDG tasks.

\secvspace
\section{Method}\label{sec:method}
\secvspace
Let ${\mathcal{D}=\{D_n}\}_{n=1}^{N}$ be a set of $N$ domains used for training (e.g., a pair of source and target domains in FUDA~\cite{kim2021cds,Yue2021}, or a set of source domains in UDG~\cite{zhang2021domain}). Each domain $D_n$ is represented by a set of \textit{unlabeled} images $\{I_n^j\}$, for clarity we will omit $j$ and denote by $I_n$ `an image' from domain $D_n$. Our goal is to train a backbone model $\mathcal{B}$ (e.g., CNN) that projects any image $I_n \in D_n$ into a $d$-dimensional representation space $\mathcal{F} \subset \mathbb{R}^d$ (shared for all domains), in a way that for a class mapping $\mathcal{C}$ (unknown at training) and any $I_n \in D_n$ and $I_m,I_r \in D_m$, s.t. $\mathcal{C}(I_n) = \mathcal{C}(I_m) \neq \mathcal{C}(I_r)$: $||\mathcal{B}(I_n) - \mathcal{B}(I_m)|| \ll ||\mathcal{B}(I_n) - \mathcal{B}(I_r)||$ will likely be satisfied. Moreover, our overall goal is that this semantic alignment property will also generalize to other domains, even if they are not seen during training. 

We train $\mathcal{B}$ using contrastive self-supervised learning extending the MocoV2 approach~\cite{chen2020improved} to incorporate \oursspace learning and other ideas explained below. Specifically, our training architecture (Figure~\ref{fig:arch}) is comprised of the following components: 
\textbf{\blue{(1)}} The backbone $\mathcal{B}:I \to \mathbb{R}^d$ (e.g., ResNet50 with GAP and $d=2048$) - it is the only thing kept after training, the rest of the components are used for training only and later discarded; 
\textbf{\blue{(2)}} The projection head $\mathcal{P}:\mathbb{R}^d \to \mathbb{R}^p$ where $p<d$ (e.g., two-layer MLP with $p=128$) and with $\mathbb{L}_2$ normalization on top;
%
\textbf{\blue{(3)}} Separate negatives queue $\mathcal{Q}_n$ for each of the domains $D_n \in \mathcal{D}$ - as separating between domains is much easier than separating between classes, we observed that having a single queue (as in \cite{chen2020improved}) for all the domains hurts performance (as explained in \cref{sec:ablations}); 
\textbf{\blue{(4)}} A set of image-to-image models $\Psi_n:D_n \to \Omega$, for mapping each domain $D_n \in \mathcal{D}$ to the shared (across all seen and unseen domains) auxiliary \oursspace domain $\Omega$ - the $\Psi_n$ are regularized to produce edge-like images which comprise $\Omega$, in \cref{sec:results} we explore and compare several options for $\Psi_n$; 
\textbf{\blue{(5)}} Domain discriminator $\mathcal{A}:\mathbb{R}^d \to \{1,\cdots,N\}$, which is an adversarial domain classifier applied \textit{only} to the $\Omega$ images representations, i.e., to $\mathcal{B}(\Psi_n(I_n))$, and trying to predict the original domain index $n$ of any image $I_n \in D_n$ projected to $\Omega$. Intuitively, learning to produce representations that confuse $\mathcal{A}$ better aligns the projections of all the different original domains inside $\Omega$.
\textbf{\blue{(6)}} The momentum models $\mathcal{B}^m$ and $\mathcal{P}^m$ (as in \cite{chen2020improved}): EMA-only updated copies of $\mathcal{B}$ and $\mathcal{P}$ respectively.

The training proceeds in batches of images randomly sampled from all the training domains $\mathcal{D}$ jointly. For clarity we will describe the training flow for a single input image $I_n \in D_n \in \mathcal{D}$. Having $I_n^{a1}$ and $I_n^{a2}$ be two augmentations of $I_n$, we first define the following contrastive loss:
\begin{align}
\begin{split}
    \mathcal{L}_{cont}(I_n) &=
    \mathcal{L}_{nce}(\mathcal{P}(\mathcal{B}(I_n^{a1})),\mathcal{P}^m(\mathcal{B}^m(\Psi_n(I_n^{a2}))),\mathcal{Q}_n)\\
    &+\mathcal{L}_{nce}(\mathcal{P}(\mathcal{B}(\Psi_n(I_n^{a1}))),\mathcal{P}^m(\mathcal{B}^m(I_n^{a2})),\mathcal{Q}_n)
\end{split}\label{eq:cont_loss}
\end{align}
where $\mathcal{L}_{nce}(q,k_+,k_-)$ is the standard InfoNCE loss \cite{gutmann2010noise,he2020momentum} with the query $q$, the positive key $k_+$ that attracts $q$, and the set of negative keys $k_-$ that repulse $q$. Our InfoNCE uses cosine similarity to compare queries and keys. Since in both $\mathcal{L}_{nce}$ summands of \cref{eq:cont_loss}, the positive keys $k_+$ are always encoded via the momentum models $\mathcal{B}^m$ and $\mathcal{P}^m$ (not producing gradients), we need both these $\mathcal{L}_{nce}$ in order to
train $\mathcal{B}$ and $\mathcal{P}$ to represent both the original training domains images $I_n \in D_n$ and their \ours-mappings $\Psi_n(I_n) \in \Omega$.
%
%
Note that the first $\mathcal{L}_{nce}$ of \cref{eq:cont_loss} teaches $\mathcal{B}$ to extract $\Omega$-relevant features directly from each $D_n$, which means we can discard the \ours-mapping models $\Psi_n$ after training and apply $\mathcal{B}$ even to unseen domains for which we do not have a learned \ours-mapping. 
After each batch is processed, the batch images `momentum' representations are (circularly) queued in accordance to their source domains: 
\begin{equation}
\mathcal{Q}_n \leftarrow \mathcal{Q}_n \cup \{ \mathcal{P}^m(\mathcal{B}^m(\Psi_n(I_n^{a2}))),
\mathcal{P}^m(\mathcal{B}^m(I_n^{a2}))
\}    
\end{equation}
Maintaining our queues in this way enables $D_n$ images from future training batches to contrast (in $\mathcal{F}$) not only against $\Omega$ projections of other $D_n$ images, but also against other images from $D_n$ - thus enabling our representation model $\mathcal{B}$ to complement its set of $\Omega$-specific features with some $D_n$-specific ones (e.g. color features).
Additionally, we use the following adversarial loss:
\begin{equation}
    \mathcal{L}_{adv}(I_n) = CE(\mathcal{A}(\mathcal{B}(\Psi_n(I_n^{a1}))),n)
    \label{eq:adv_loss}
\end{equation}
where $CE$ is the standard cross-entropy loss and $n \in \{1,\cdots,N\}$ is the correct domain index for the image $I_n$. We employ the standard (`two-optimizers' in PyTorch) adversarial training scheme for $\mathcal{L}_{adv}$. In each training batch, the domain discriminator $\mathcal{A}$ is minimizing $\mathcal{L}_{adv}$, while blocking $\mathcal{B}$ and $\Psi_n$ gradients, whereas $\mathcal{B}$ and $\Psi_n$ minimize the negative loss: $-\mathcal{L}_{adv}$, while blocking the $\mathcal{A}$ gradients. 
Note that we employ $\mathcal{L}_{adv}$ only for the $\Omega$ projections of the original domains, thus not requiring direct alignment between the domains of $\mathcal{D}$. Moreover, to reduce `competition' between $\mathcal{L}_{adv}$ and $\mathcal{L}_{cont}$, we use the domain discriminator $\mathcal{A}$ directly on $\mathcal{B}$-generated representations (the final features) and not on the projection head $\mathcal{P}$-generated representations (temporary features used for efficiency in $\mathcal{L}_{cont}$).
Lastly, we define the \oursspace loss that regularizes the $\Psi_n$ models to produce edge-like images comprising the shared auxiliary \oursspace domain $\Omega$, which we show to be very effective for different tasks in \cref{sec:results}:
\begin{equation}
    \mathcal{L}_{\Omega}(I_n) = ||\Psi_n(I_n^{a1}) - \mathcal{E}(I_n^{a1})||^2
    \label{eq:loss_omega}
\end{equation}
where $\mathcal{E}$ is some edge model, which could be heuristic, such as Canny edge detector~\cite{canny1986computational}, or pre-trained, such as HED~\cite{Xie_2015_ICCV}, we explore and compare variants in \cref{sec:ablations}.
Finally, our full loss for image $I_n$ is therefore:
\begin{equation}
    \mathcal{L}_{f}(I_n) = 
    \alpha_1 \cdot \mathcal{L}_{cont}(I_n)
    + \alpha_2 \cdot \mathcal{L}_{\Omega}(I_n)
    - \alpha_3 \cdot \mathcal{L}_{adv}(I_n)
\end{equation}
where the sign in front of $\mathcal{L}_{adv}$ becomes positive when computing gradients for training the adversarial domain discriminator $\mathcal{A}$.

\noindent\textbf{Implementation details.} Our code\footnote{Available at \url{https://github.com/leokarlin/BrAD}} is in PyTorch~\cite{torch} and is based on the code of~\cite{chen2020improved}. 
We set $\alpha_1,\alpha_2,\alpha_3=1$ in our experiments. The backbone $\mathcal{B}$ was ResNet-18~\cite{He_2016_CVPR} for UDG experiments (same as in~\cite{zhang2021domain}), and ResNet-50 in FUDA and cross-benchmark generalization experiments (same as in~\cite{kim2021cds}). We used batches of size $256$, SGD with momentum $0.9$, cosine LR-schedule (from LR 0.03 to 0.002) and trained for $250$ epochs for FUDA and $1000$ epochs for UDG (same as \cite{zhang2021domain}). We set $|\mathcal{Q}_n|=min(64K,2\cdot|D_n|)$ and stored only a single pair of (momentum) representations generated from each domain image $I_n$ and its $\Omega$ projection (by $\Psi_n$). 
Furthermore, we found it slightly beneficial to exclude the cached version of $q$ from its $k_-$ negative keys set when computing the $\mathcal{L}_{nce}(q,k_+,k_-)$ losses. 
For $\mathcal{A}$ we used a $3$-layer MLP($1024, 512, 256$) with LeakyReLU, followed by a linear domain classifier. For the \ours-mapping models $\Psi_n$ architecture, we used the architecture of HED~\cite{Xie_2015_ICCV} in its PyTorch implementation~\cite{pytorch-hed}.

\begin{table*}[t]
    \vspace{-0.5cm}
    \centering
    \small
    \begin{tabular}{lcccccc|cc}
        \toprule
         Source domains & \multicolumn{3}{c}{\{Paint. $\cup$ Real $\cup$ Sketch\}} & \multicolumn{3}{c}{\{Clipart $\cup$ Info. $\cup$ Quick.\}}\\
         Target domain & Clipart & Info. & Quick. & Painting & Real & Sketch & Overall & Avg. \\
        \midrule

         \multicolumn{9}{c}{Label Fraction 1\%} \\
         \midrule
        ERM  & 6.54 & 2.96 & 5.00 & 6.68 & 6.97 & 7.25 & 5.88 & 5.89  \\
        BYOL \cite{grill2020bootstrap} & 6.21 & 3.48 & 4.27 & 5.00 & 8.47 & 4.42 & 5.61 & 5.30  \\
        MoCo V2 \cite{chen2020improved, he2020momentum}  & 18.85 & 10.57 & 6.32 & 11.38 & 14.97 & 15.28 & 12.12  & 12.90 \\
        AdCo \cite{hu2021adco}  & 16.16 & 12.26 & 5.65 & 11.13 & 16.53 & 17.19 & 12.47 & 13.15  \\
        SimCLR V2 \cite{chen2020big}  & {23.51} & \textcolor{blue}{15.42} & 5.29 & \textbf{20.25} & 17.84 & 18.85 & 15.46 & 16.55  \\
        DIUL \cite{zhang2021domain}  & 18.53 & 10.62 & {12.65} & 14.45 & {21.68} & {21.30} & {16.56} & 16.53  \\
        \midrule
        Ours (kNN) & \textcolor{blue}{40.65} & {14.00} & \textcolor{blue}{21.28} & 16.80 & \textcolor{blue}{22.29} & \textcolor{blue}{25.72} & \textcolor{blue}{22.35} & \textcolor{blue}{23.46}  \\
        Ours (linear cls.) & \textbf{47.26} & \textbf{16.89} & \textbf{23.74} & \textcolor{blue}{20.03} & \textbf{25.08} & \textbf{31.67} & \textbf{25.85} & \textbf{27.45} \\
        \midrule
        
        \multicolumn{9}{c}{Label Fraction 5\%} \\
         \midrule
        ERM   & 10.21 & 7.08 & 5.34 & 7.45 & 6.08 & 5.00 & 6.50 & 6.86  \\
        BYOL \cite{grill2020bootstrap} & 9.60 & 5.09 & 6.02 & 9.78 & 10.73 & 3.97  & 7.83 & 7.53 \\
        MoCo V2 \cite{chen2020improved, he2020momentum} & 28.13 & 13.79 & 9.67 & 20.80 & 24.91 & 21.44 & 18.99 & 19.79 \\
        AdCo \cite{hu2021adco}  & 30.77 & 18.65 & 7.75 & 19.97 & 24.31 & 24.19 & 19.42 & 20.94 \\
        SimCLR V2 \cite{chen2020big}  & 34.03 & 17.17 & {10.88} & 21.35 & 24.34 & 27.46  & 20.89 & 22.54 \\
        DIUL \cite{zhang2021domain} &  39.32 & \textcolor{blue}{19.09} & 10.50 & 21.09 & 30.51 & 28.49 & 23.31 & 24.83 \\
        \midrule
        Ours (kNN) & \textcolor{blue}{55.75} & 18.15 & \textcolor{blue}{26.93} & \textcolor{blue}{24.29} & \textcolor{blue}{33.33} & \textcolor{blue}{37.54} & \textcolor{blue}{31.12} & \textcolor{blue}{32.66} \\
        Ours (linear cls.) & \textbf{64.01} & \textbf{25.02} & \textbf{29.64} & \textbf{29.32} & \textbf{34.95} & \textbf{44.09} & \textbf{35.37} & \textbf{37.84} \\
        \midrule
        
        \multicolumn{9}{c}{Label Fraction 10\%} \\
         \midrule
        ERM  & 15.10 & 9.39 & 7.11 & 9.90 & 9.19 & 5.12 & 8.94 & 9.30  \\
        BYOL \cite{grill2020bootstrap} & 14.55 & 8.71 & 5.95 & 9.50 & 10.38 & 4.45 & 8.69 & 8.92  \\
        MoCo V2 \cite{chen2020improved, he2020momentum} & 32.46 & 18.54 & 8.05 & 25.35 & 29.91 & 23.71 & 21.87 & 23.05 \\
        AdCo \cite{hu2021adco} & 32.25 & 17.96 & 11.56 & 23.35 & 29.98 & 27.57 & 22.79 & 23.78  \\ 
        SimCLR V2 \cite{chen2020big} & {37.11} & 19.87 & 12.33 & 24.01 & 30.17 & 31.58 & 24.28 & 25.84 \\
        DIUL \cite{zhang2021domain} & 35.15 & \textcolor{blue}{20.88} & {15.69} & {25.90} & {33.29} & 30.77 & {26.09} & 26.95  \\
        \midrule
        Ours (kNN) & \textcolor{blue}{60.78} & {19.76} & \textcolor{blue}{31.56} & \textcolor{blue}{26.06} & \textcolor{blue}{37.43} & \textcolor{blue}{41.38} & \textcolor{blue}{34.77} & \textcolor{blue}{36.16}  \\
        Ours (linear cls.) & \textbf{68.27} & \textbf{26.60} & \textbf{34.03} & \textbf{31.08} & \textbf{38.48} & \textbf{48.17} & \textbf{38.74} & \textbf{41.10} \\
        
        \bottomrule
    \end{tabular}
    \caption{Accuracy (\%) results for UDG on DomainNet. All baseline results are taken from \cite{zhang2021domain}. All methods use ResNet18 backbone and are unsupervisedly pretrained for 1000 epoches before training on the labeled (source only) data. All baselines use a linear classifier (for ours we also include a kNN result that does not utilize any supervised training). ERM indicates the randomly initialized ResNet18. Overall and Avg. indicate the overall test data accuracy and the mean of the per-domain accuracies, respectively. They are different since the test sets of different domains are not of the same size. The reported results are an average over $3$ runs. 
     \textbf{bold} = best, \textcolor{blue}{blue} = second best.}
    \label{tab:all_correlated_domainet}
    \tabvspace
\end{table*}
\secvspace
\section{Results}\label{sec:results}
\secvspace
As our \oursspace approach is completely unsupervised during training, we used none or limited-supervision cross-domain tasks, specifically Unsupervised Domain Generalization (UDG)~\cite{zhang2021domain} and the Few-shot UDA (FUDA)~\cite{kim2021cds,Yue2021}, for evaluating its performance and comparing to other self-supervised or source-labels-limited cross-domain methods. Additionally, we evaluate how \oursspace and other self-supervised approaches generalize to unseen domains and unseen classes after being trained on a large-scale unlabeled cross-domain data, such as DomainNet~\cite{peng2019moment}.

\noindent\textbf{Datasets.} 
\textbf{\blue{DomainNet}}~\cite{peng2019moment} is the largest, most diverse and recent cross-domain benchmark to-date. It is comprised of $6$ domains: Real, Painting, Sketch, Clipart, Infograph and Quickdraw, with $345$ object classes, 48K - 173K images per domain, and average of $269$ images per class.
\textbf{\blue{PACS}}~\cite{Li_2017_ICCV} is a standard domain generalization benchmark. It is comprised of $4$ domains: Photo, Art, Cartoon and Sketch, with $7$ object classes, 2.5K images per domain, and average of $357$ images per class.
\textbf{\blue{VisDA}}~\cite{visda2017} is a simulation-to-real dataset with $12$ classes. The simulation domain is generated via repeated (80-480 times) 3D rendering of instances of 3D object models, 50-150 models per class. It is therefore comprised of only $\sim$1.5K distinct object instances.
%
\textbf{\blue{Office-Home}}~\cite{venkateswara2017deep} is a relatively small dataset consisting of $4$ domains: Art, Clipart, Product and Real, with $65$ classes, and an average of only $60$ images per class.

\secvspace
\subsection{Unsupervised Domain Generalization}\label{sec:UDG}
\secvspace
The UDG task is defined as: \textbf{\blue{(i)}} unsupervised training on a set of source domains; \textbf{\blue{(ii)}} using only a small labeled subset of source domain images to fit a linear classifier on top of the (frozen) features produced by the unsupervised model; and \textbf{\blue{(iii)}} evaluating the resulting classifier performance on a set of target domains, unseen during training. In our UDG experiments we accurately followed the protocol of the UDG state-of-the-art (SOTA) method DIUL~\cite{zhang2021domain}, including same backbone arch., same num. epochs, and same subset of classes used for training and testing. Same as \cite{zhang2021domain}, we evaluated on DomainNet~\cite{peng2019moment} (\cref{tab:all_correlated_domainet}) and PACS~\cite{Li_2017_ICCV} (\cref{tab:all_correlated_pacs}). In DomainNet, we train on Clipart, Infograph and Quickdraw and test on unseen Painting, Real and Sketch, and vice versa. For PACS we do a leave-one-domain-out test using the other three as source (repeating this for all domains). 
%
Unlike DIUL~\cite{zhang2021domain}, who used additional full model fine-tuning when amount of source labels was $10\%$ of the source data size, our self-supervised model was \textit{never} fine-tuned with the source labels (in all cases). Moreover, we also provide kNN results for our method, where we use our resulting features directly without any additional training. 

As can be seen from \cref{tab:all_correlated_domainet} and \cref{tab:all_correlated_pacs}, \oursspace demonstrates significant gains (both in linear cls. and in kNN modes) not only over~\cite{zhang2021domain}, but also over a variety of SOTA self-supervised pre-training baselines (probed with the classifier in exactly the same manner as \cite{zhang2021domain} and as described above). This illustrates an important point, the \oursspace idea seems to be effective in improving the generalization of self-supervised pre-training to unseen target domains, which, according to these results, seems quite difficult for the current self-supervised SOTA methods.
\begin{table}[t]
    \centering
    \resizebox{1\linewidth}{!}{
    \begin{tabular}{lcccc|c}
        \toprule
         Target domain & Photo & Art. & Cartoon & Sketch & Avg.\\
        \midrule
         \multicolumn{6}{c}{Label Fraction 1\%} \\
         \midrule
        ERM  & 10.90 & 11.21 & 14.33 & 18.83 & 13.82 \\
        BYOL \cite{grill2020bootstrap} & 11.20 & 14.53 & 16.21 & 10.01 & 12.99  \\
        MoCo V2 \cite{chen2020improved, he2020momentum} & 22.97 & 15.58 & 23.65 & 25.27 & 21.87  \\
        AdCo \cite{hu2021adco} & 26.13 & 17.11 & 22.96 & 23.37 & 22.39 \\
        SimCLR V2 \cite{chen2020big} & {30.94} & 17.43 & {30.16} & 25.20 & 25.93  \\
        DIUL \cite{zhang2021domain} & 27.78 & {19.82} & 27.51 & {29.54} & {26.16} \\
        \midrule
        Ours (kNN) & \textcolor{blue}{55.00} & \textbf{35.54} & \textcolor{blue}{38.12} & \textcolor{blue}{34.14} & \textcolor{blue}{40.70}  \\
        Ours (linear cls.) & \textbf{61.81} & \textcolor{blue}{33.57} & \textbf{43.47} & \textbf{36.37} & \textbf{43.81} \\
        \midrule
        
         \multicolumn{6}{c}{Label Fraction 5\%} \\
         \midrule
        ERM   & 14.15 & 18.67 & 13.37 & 18.34 & 16.13 \\
        BYOL \cite{grill2020bootstrap} & 26.55 & 17.79 & 21.87 & 19.65 & 21.47 \\
        MoCo V2 \cite{chen2020improved, he2020momentum} & 37.39 & 25.57 & 28.11 & 31.16 & 30.56 \\
        AdCo \cite{hu2021adco} & 37.65 & 28.21 & 28.52 & 30.35 & 31.18 \\
        SimCLR V2 \cite{chen2020big} & {54.67} & 35.92 & 35.31 & {36.84} & {40.68} \\
        DIUL \cite{zhang2021domain} & 44.61 & \textcolor{blue}{39.25} & {36.41} & 36.53 & 39.20 \\
        \midrule
        Ours (kNN) & \textcolor{blue}{58.66} & 39.11 & \textcolor{blue}{45.37} & \textcolor{blue}{46.11} & \textcolor{blue}{47.31} \\
        Ours (linear cls.) & \textbf{65.22} & \textbf{41.35} & \textbf{50.88} & \textbf{50.68} & \textbf{52.03} \\
        
        \midrule
        \multicolumn{6}{c}{Label Fraction 10\%} \\
         \midrule
     
        ERM  & 16.27 & 16.62 & 18.40 & 12.01 & 15.82  \\
        BYOL \cite{grill2020bootstrap} & 27.01 & 25.94 & 20.98 & 19.69 & 23.40 \\
        MoCo V2 \cite{chen2020improved, he2020momentum} & 44.19 & 25.85 & 33.53 & 24.97 & 32.14 \\
        AdCo \cite{hu2021adco} & 46.51 & 30.21 & 31.45 & 22.96 & 32.78  \\ 
        SimCLR V2 \cite{chen2020big} & {54.65} & 37.65 & 46.00 & 28.25 & 41.64 \\
        DIUL \cite{zhang2021domain} & 53.37 & {39.91} & \textcolor{blue}{{46.41}} & {30.17} & {42.47} \\
        \midrule
        Ours (kNN) & \textcolor{blue}{67.20} & \textcolor{blue}{41.99} & 45.32 & \textcolor{blue}{50.04} & \textcolor{blue}{51.14}  \\
        Ours (linear cls.) & \textbf{72.17} & \textbf{44.20} & \textbf{50.01} & \textbf{55.66} & \textbf{55.51} \\
        \bottomrule
    \end{tabular}}
    
    \caption{Accuracy (\%) results for the UDG on PACS. For each target domain all other 3 are used as source domains for training. For other details about the number of runs, meaning of column titles and etc., please see \cref{tab:all_correlated_domainet} caption. All baseline results are taken from \cite{zhang2021domain}. \textbf{bold} = best results, \textcolor{blue}{blue} = second best.
    }
    \label{tab:all_correlated_pacs}
    \tabvspace
\end{table}

\secvspace
\subsection{Few-shot Unsupervised Domain Adaptation} \label{sec:FSUDA}
\secvspace
We used the largest and most recent cross-domain dataset, DomainNet~\cite{peng2019moment}, to evaluate our \oursspace approach FUDA~\cite{kim2021cds,Yue2021} performance. Same as in~\cite{Yue2021} (and as is common UDA practice), for this evaluation we used only $4$ domains: Clipart, Real, Painting and Sketch, and the source-target directions listed in \cref{tab:FSUDA_results}. We follow the FUDA protocol defined in~\cite{Yue2021}, where source domain has a single ($1$-shot) or three ($3$-shot) labeled images per-class and the rest of its images are provided as unlabeled.
We used the same classes and the exact indices of the labeled samples for each case as provided by~\cite{Yue2021} for repeatability.
Our results and comparison to other methods are summarized in \cref{tab:FSUDA_results}. 
According to the protocol of~\cite{Yue2021}, all compared methods models are initialized with ImageNet pre-training and operate in transductive setting\footnote{According to the released official code of~\cite{Yue2021}, \blue{transductive} = utilize the entire domains data (including unlabeled test data) in their training. Transductive setting adds about $3-4\%$ to the performance (\cref{sec:ablations}).}. All the methods besides ours use the respective $1$ or $3$ samples per class during their training. In our case, we used those samples \textit{only during the inference} - either as the search space in the kNN, or for training the linear classifier (on these few labeled source images only). All the methods besides ours are designed for working separately on each pair of source and target domains. Therefore, for the comparison to be comprehensive, we include results for our method in both our intended mode and a \textit{pairwise} mode. In our intended mode (`ours' in \cref{tab:FSUDA_results}) we train a single model on all the domains jointly. In the \textit{pairwise} mode (`ours pairwise' in \cref{tab:FSUDA_results}), we train $7$ separate models, one for each source-target domain pair.
%
%
Besides showing a competitive advantage of our method in all modes, results in \cref{tab:FSUDA_results} indicate that multi-domain training has clear advantages in efficiency (single model vs. $7$ models), ease of use (no need to know the query domain), and performance (about $10\%$ better).
%

\begin{table*}[t]
    \vspace{-0.5cm}
    \centering
    
    \resizebox{1\textwidth}{!}{
    \begin{tabular}{lccccccc|c}
        \toprule
         Source domain & Real & Real & Real & Painting & Painting & Clipart & Sketch & \\
         Target domain & Clipart & Painting & Sketch & Clipart & Real & Sketch & Painting & Avg. \\
        \midrule
Source-Only baseline \cite{Yue2021} & 18.4 / 30.2 & 30.6 / 44.2 & 16.7 / 25.7 & 16.2 / 24.6 & 28.9 /  49.8  &12.7 / 24.2 & 10.5 / 23.2 & 19.1 / 31.7 \\
MME\cite{Saito19} & 13.8 / 22.8 & 29.2 / 46.5 & 9.7 / 14.5 & 16.0 / 25.1 &  26.0 / 50.0 & 13.4 / 20.1 & 14.4 / 24.9 & 17.5 / 29.1 \\
CDAN \cite{Long18} & 16.0 / 30.0 & 25.7 / 40.1 & 12.9 / 21.7 & 12.6 / 21.4 & 19.5 / 40.8 & 7.20 / 17.1 & 8.00 / 19.7 & 14.6 / 27.3 \\
MDDIA \cite{Jiang20} & 18.0 / 41.4 & 30.6 / 50.7 & 15.9 / 37.4 & 15.4 / 31.4 & 27.4 / 52.9 & 9.30 / 23.1 & 10.2 / 24.1 & 18.1 / 37.3 \\
CAN \cite{Kang19} & 18.3 / 28.1 & 22.1 / 33.5 & 16.7 / 25.0 & 13.2 / 24.7 & 23.9 / 46.9 & 11.1 / 23.3 & 12.1 / 20.1 & 16.8 / 28.8 \\ 
CDS+SRDC+ENT \cite{kim2021cds} & 23.1 / 36.6 & 40.0 / 54.0 & 22.2 / 35.5 & 24.1 / 38.1 & 35.1 / 57.6 & 18.8 / 35.4 & 25.2 / 45.1 & 26.9 / 43.2\\ 
CDS+MME+ENT \cite{kim2021cds} & 35.4 / 47.4 & 36.7 / 52.8 & 33.4 / 43.2 & 25.6 / 41.2 & 29.4 / 56.4 & 19.3 / 37.5 & 22.5 / 41.1 & 28.9 / 45.6\\
PCS \cite{Yue2021} & {39.0} / 45.2 & {51.7} / {59.1} & {39.8} / {41.9} & {26.4} / {41.0} & {38.8} / \textbf{66.6} & {23.7} / 31.9 & {23.6} / 37.4 & {34.7} / {46.1} \\ 
\midrule
Ours pairwise (kNN) & 43.6 / 49.5 & 50.4 / 57.0 & 41.7 / 47.9 & 35.9 / 41.0 & 44.2 / 60.7 & 34.5 / 42.4 & 36.1 / 45.5 & 40.9 / 49.1 \\
Ours pairwise (linear cls.) & 44.0 / 51.4 & 50.1 / 58.9 & 47.0 / 55.6 & 35.7 / 42.5 & 44.5 / 62.1 & 35.3 / 45.0 & 35.65 / 45.0 & 41.8 / 51.5 \\
\midrule
Ours (kNN) & \textcolor{blue}{46.4} / \textcolor{blue}{57.6} & \textcolor{blue}{52.0} / \textcolor{blue}{59.4} & \textcolor{blue}{50.5} / \textcolor{blue}{58.9} & \textcolor{blue}{45.1} / \textcolor{blue}{55.3} & \textbf{49.3} / {61.3} & \textbf{48.3} / \textcolor{blue}{56.4} & \textcolor{blue}{49.0} / \textcolor{blue}{59.3} & \textcolor{blue}{48.6} / \textcolor{blue}{58.3} \\
Ours (linear cls.) & \textbf{48.6} / \textbf{60.6} & \textbf{55.1} / \textbf{62.8} & \textbf{52.8} / \textbf{61.6} & \textbf{44.6} / \textbf{56.6} & \textcolor{blue}{47.8} / \textcolor{blue}{63.6} & \textcolor{blue}{47.9} / \textbf{59.8} & \textbf{51.0} / \textbf{61.0} & \textbf{49.7} / \textbf{60.8} \\       
        \bottomrule
        
    \end{tabular}
    }
    \caption{1-shot/3-shot accuracy (\%) results for FUDA task \cite{Yue2021} on DomianNet. All baseline results except CDS \cite{kim2021cds} are taken from \cite{Yue2021}. The CDS results were kindly provided by its authors and are higher than those reported in \cite{Yue2021}. \textbf{bold} = best results, \textcolor{blue}{blue} = second best.}
    \label{tab:FSUDA_results}
    \tabvspace
\end{table*}

\secvspace
\subsection{Generalization to unseen domains and classes}
\secvspace
One of the interesting research questions to consider w.r.t. self-supervised learning - is how well it can generalize to unseen domains and classes. Assume one had access to a large collection of diverse unlabeled multi-domain data (e.g. unlabeled DomainNet~\cite{peng2019moment}) needed for training a contrastive self-supervised model\footnote{One of the known limitations of self-supervised contrastive learning methods (including ours) is the need to observe relatively large amount of \textit{different} instances per class (naturally without class labels).}. However, also assume that the trained model would need to be used for inference in a few-shot scenario (i.e. with very little data) in a new unseen domain and for a new unseen set of classes. 
As an example, consider the situation we discussed in the introduction: our self-supervised system (pre-trained on some multi-domain data) would need to recognize in a real photo an instance of unseen before technical equipment after seeing it for the first time illustrated using some proprietary graphical style in some technical manual.

To test how leading self-supervised methods~\cite{chen2020improved,caron2021emerging,caron2020unsupervised,chen2021exploring,zbontar2021barlow}
deal with the proposed scenario (generalization to a mix of seen and unseen domains, as well as mostly unseen classes) and compare their performance to our approach, we conducted the following cross-dataset FUDA generalization experiment, whose results are provided in \cref{tab:cross_dataset_ssl}. We train all methods using their official codes and recommended hyper-parameter and backbone settings on the entire data of Clipart, Real, Painting and Sketch domains from DomainNet. Same ResNet50 backbone is used for all methods (including ours) except Dino \cite{caron2021emerging} that uses the stronger ViT backbone for which it was optimized. We then evaluated the resulting models using a kNN classifier and the FUDA setting detailed in \cref{sec:FSUDA}, on the OfficeHome~\cite{venkateswara2017deep}, PACS~\cite{Li_2017_ICCV}, and VisDA~\cite{visda2017} cross-domain datasets. In all cases, the $1$-shot and the $3$-shot source domain examples per class were sampled randomly and were kept the same for all methods. This experiment (sampling of shots) was repeated $5$ times and in \cref{tab:cross_dataset_ssl} we report the averages. Similar to UDG experiments in \cref{sec:UDG}, these results indicate again the difficulty for cross-domain generalization inherent to the popular self-supervised learning approaches, and the advantages of \oursspace for improving this generalization.
In \cref{apend:demo} we provide many qualitative examples of cross-domain kNN results for the PACS datset, obtained using kNN in the features space of our self-supervised \oursspace model trained on DomainNet.

\begin{table}[b]
    \tabvspace
    \centering
    \resizebox{1\linewidth}{!}{
    \begin{tabular}{lccc}
        \toprule
          Method & OfficeHome& PACS & VisDA \\
        \midrule
        Dino \cite{caron2021emerging} & 12.41 / 16.97 & 30.90 / 34.67 & 25.02 / 28.14\\
        SWAV\cite{caron2020unsupervised} & 13.26 / 17.80 & 31.14 / 33.09 & 25.65 / 29.26\\
        SimSiam \cite{chen2021exploring} & 13.67 / 18.27 & 30.24 / 32.27 & 24.70 / 28.80\\
        BarlowTwins\cite{zbontar2021barlow} & \textcolor{blue}{17.86} / \textcolor{blue}{24.06}  & 41.18 /  46.18 & 25.34 / 30.47\\
        MocoV2~\cite{chen2020improved} & 17.64 / 22.63 & \textcolor{blue}{49.00} /   \textcolor{blue}{54.25}& \textcolor{blue}{30.34} /	 \textcolor{blue}{36.10}\\
        \midrule
        Ours &\textbf{21.79} / \textbf{28.21}& \textbf{55.61} / \textbf{63.00} & \textbf{32.98} / \textbf{40.22} \\
        \bottomrule
    \end{tabular}
    }
    \caption{1-shot / 3-shot accuracy (\%) cross dataset results. The models are trained from scratch on DomainNet and tested on the OfficeHome, PACS and VisDA. 
    We report an average over $5$ runs randomizing the shots.
    \textbf{bold} = best, \textcolor{blue}{blue} = second best. }
    \label{tab:cross_dataset_ssl}
\end{table}

\subsection{Ablation Studies}\label{sec:ablations}
\secvspace
In \cref{tab:ablation_main} we evaluate the contribution of the different components of our approach using the FUDA task~\cite{kim2021cds,Yue2021} on the DomainNet dataset~\cite{peng2019moment}. The experimental setting is described in \cref{sec:FSUDA} above.
Specifically, we show how the average performance of the resulting model in $1$-shot / $3$-shots FUDA on DomainNet evolves starting from the vanilla MocoV2~\cite{chen2020improved} and adding:
\textbf{\blue{(i)}} DD: a domain discriminator ($\mathcal{A}$ in \cref{eq:adv_loss}) - on its own it has a minor impact on performance ($\sminus0.3/\splus0.1$); 
\textbf{\blue{(ii)}} MQ: multiple negative queues ($\mathcal{Q}_n$) for contrastive loss - adds good boost to $1$-shot case ($\splus4.2/\splus0.4$) on its own, and strong boost for both modes when combined with DD ($\splus3.8/\splus6.0$); 
\textbf{\blue{(iii)}} Canny \ours: $\Psi_n$ in heuristic Canny~\cite{canny1986computational}  edge detector form - leads to a very strong performance boost ($\splus10.9/\splus11.9$) underlining the effectiveness of the \oursspace idea; 
\textbf{\blue{(iv)}} HED \ours: $\Psi_n$ being a frozen HED~\cite{Xie_2015_ICCV} edge detector pre-trained on BSDS500~\cite{amfm_pami2011} dataset  - we observe that even using a strong pre-trained edge detector model is not sufficient to further improve relative to the simpler Canny \oursspace ($\sminus1.7/\sminus1.8$), this clearly highlights that \oursspace models $\Psi_n$ need to be learned jointly (end-to-end) with the representation model $\mathcal{B}$ as we propose in our main approach;
\textbf{\blue{(v)}} learned \ours: $\Psi_n$ being a HED~\cite{Xie_2015_ICCV} model trained end-to-end with the other components of our \oursspace approach as described in \cref{sec:method} - underlining the need to \textit{learn} the \oursspace $\Psi_n$ models, this introduces a noticeable boost relative to the heuristic Canny \oursspace ($\splus2.8/\splus2.3$) and overall compared to not using \oursspace ($\splus13.7/\splus14.2$); 
\textbf{\blue{(vi)}} Typical examples of comparison of edges generated by Canny, pretrained HED \cite{Xie_2015_ICCV}, and our learned \oursspace are shown in Fig. \ref{fig:edges_1col} - as can be seen, both \oursspace and HED discard the background noise, but unlike HED, \oursspace learns to retain semantic details of shape and texture like house windows, giraffe spots, or person arm (additional examples are provided in \cref{apend:edge_maps}); 
\textbf{\blue{(vii)}} Transductive / ImageNet pretrained: according to the FUDA experimental setting of PCS~\cite{Yue2021}, used for all methods in our FUDA evaluation in \cref{sec:FSUDA}, training starts from an ImageNet pretrained model and transductive paradigm is used for the unlabeled domains data - we have verified that the transductive setting consistently adds $\sim 4\%$ regardless of pretraining, while ImageNet pretraining has a more significant impact, adding $\sim 10\%$ to the performance.

%
        


\begin{table}[t]
    \centering
    \resizebox{1\linewidth}{!}{
    \begin{tabular}{ccccccc}
        \toprule
         & & \ours  & Transd- & ImageNet & & \\
         DD & MQ & $\Psi_n$ & uctive&pretrained  &1-shot  & 3-shots \\
        \midrule
        - & - & - & - & - & 16.8 &  21.9\\
        \checkmark & - & - & - & - & 16.5 &  22.0\\
        - & \checkmark & - & - & - & 21.0 &  22.3\\
        \checkmark & \checkmark & - & - & - & 20.6 & 27.9\\
        \checkmark & \checkmark  & Canny & - & - & 31.5 & 39.8\\
        \checkmark & \checkmark & HED & - & - & 29.8 & 38.0\\
        \checkmark & \checkmark & learned & - & - & 34.3 & 42.1\\
        \checkmark & \checkmark & learned & \checkmark & - & 37.8 & 47.1\\
        \checkmark & \checkmark & learned & - & \checkmark & 44.1 & 55.2\\
        \cellcolor{blue!15}\textbf{\checkmark} & \cellcolor{blue!15}\textbf{\checkmark} & \cellcolor{blue!15}{learned} & \cellcolor{blue!15}{\checkmark} & \cellcolor{blue!15}{\checkmark} & \cellcolor{blue!15}\textbf{48.6} & \cellcolor{blue!15}{\textcolor{blue}{58.3}} \\
        \checkmark & \checkmark & ~~learned\textbf{*}  & \checkmark & \checkmark & \textcolor{blue}{47.4} & \textbf{59.3}\\
        \checkmark & \checkmark & -  & \checkmark & \checkmark & 38.7 & 51.1\\
        \checkmark & \checkmark & Canny  & \checkmark & \checkmark & 41.4 & 52.7\\
        \checkmark & \checkmark & HED  & \checkmark & \checkmark & 41.6 & 51.9\\
        \bottomrule
        
    \end{tabular}
    }
    \caption{\oursspace ablation study using the FUDA task on DomainNet dataset with knn classifier. \textbf{bold} = best, \textcolor{blue}{blue} = second best. The \colorbox{blue!15}{highlighted} row is our full method in \cref{sec:FSUDA} settings. The ``learned\textbf{*}'' is our full method without using HED pretrained weights.
    }
    \label{tab:ablation_main}
    \tabvspace
\end{table}

        

%
Summarily adding all the above components and aligning to the experimental setting of PCS~\cite{Yue2021}, we arrive at our main FUDA result (highlighted in \cref{tab:ablation_main}). Moreover, we verified that our approach does not have a strong dependency on BSDS500~\cite{amfm_pami2011} dataset pre-training of the HED models~\cite{Xie_2015_ICCV} used as $\Psi_n$. Specifically, we randomly initialized the $\Psi_n$ \oursspace models (that have HED architecture), and replaced the BSDS500 pretrained HED model used as $\mathcal{E}$ in the $\mathcal{L}_{\Omega}$ loss (\cref{eq:loss_omega}) with a simple blurred Canny edge map (more details in the Supplementary). This way \textit{not having} any BSDS500 pretrained weights anywhere in our system. As can be seen from the corresponding ``learned\textbf{*}'' row in Table~\ref{tab:ablation_main}, there is almost no change in the end result ($\sminus1.2/\splus1.0$)) indicating our approach can work just as well without BSDS500 pretraining of HED. Please refer to \cref{apend:wo_pre_hed} for more details.
Finally, we verified that the strong gains we observed due to the introduction of \oursspace models $\Psi_n$ do not disappear with ImageNet pretriaining and transductive setting. As can be seen in the corresponding rows of \cref{tab:ablation_main}, both Canny and frozen HED \oursspace variants maintain moderate gains of up to $2.9\%$ (over no-\oursspace mode), while our full method with the learned $\Psi_n$ \oursspace maintains large gains ($\splus9.9/\splus7.2$) as expected.

\newcommand*\figsizecol{0.3\columnwidth}
\newcommand*\trimall{0.5}

\begin{figure}[t]
\vspace{-0.3cm}
    \centering
    ~~~~~Real~~~~~~~~~~~~~~~~~~Art~~~~~~~~~~~~~~~~~~~~Sketch\\
    \rotatebox{90}{~~~~~Original}~\includegraphics[width=\figsizecol]{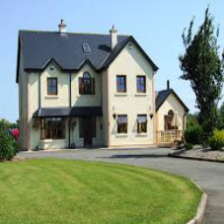}
    \includegraphics[width=\figsizecol]{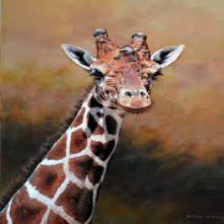}
    \includegraphics[width=\figsizecol]{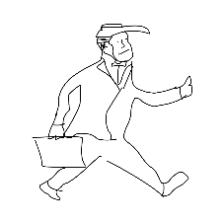}\\
    \rotatebox{90}{~~~~~~Canny \cite{canny1986computational}}~\includegraphics[width=\figsizecol]{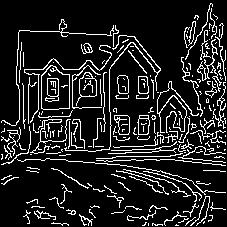}
    \includegraphics[width=\figsizecol]{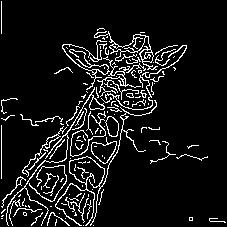}
    \includegraphics[width=\figsizecol]{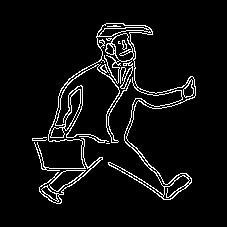}\\
    \rotatebox{90}{~~~~~HED~Zoom~~~~~~~~~~~~~HED\cite{Xie_2015_ICCV}}~\includegraphics[width=\figsizecol]{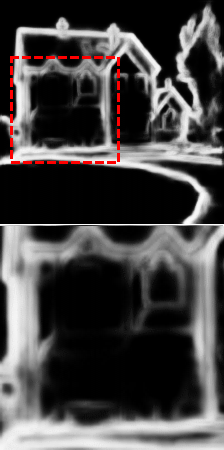}
    \includegraphics[width=\figsizecol]{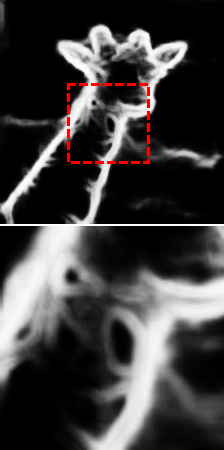}
    \includegraphics[width=\figsizecol]{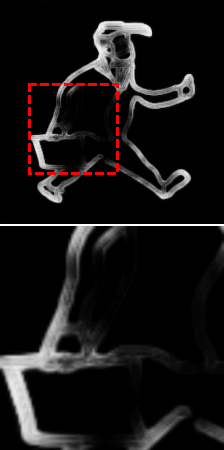}\\
    \hspace{4.5pt}\rotatebox{90}{~~~\ours{}~Zoom~~~~~~~~~~~~~~~~~\ours}~\includegraphics[width=\figsizecol]{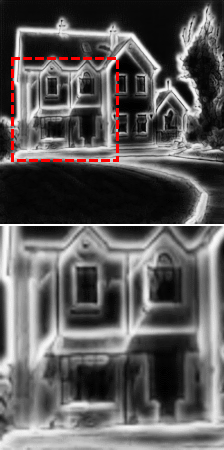}
    \includegraphics[width=\figsizecol]{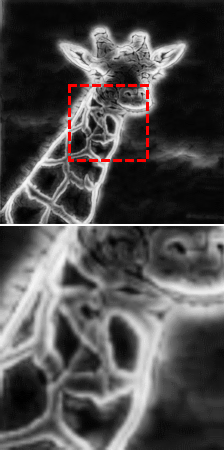}
    \includegraphics[width=\figsizecol]{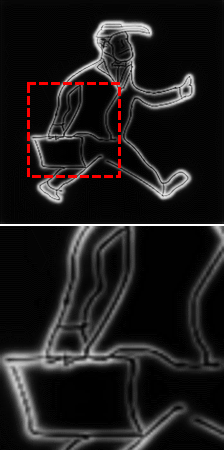}
    \vspace{-0.2cm}
    \caption{Edge images for different choices of $\Psi_n$: (i) Canny produces noisy images with many irrelevant edges; (ii) the HED mostly outlines the object and discards important internal texture; (iii) learned $\Psi_n$ keep the informative edges and the fine internal details while discarding most of the noise. More examples can be found in supplementary.}
    \label{fig:edges_1col}
    \figvspace
\end{figure}

\secvspace
\section{Conclusions and Limitations}
\secvspace
In this paper, we have proposed a novel self-supervised cross-domain learning method based on semantically aligning (in feature space) all the domains to a common \oursspace domain - a learned auxiliary bridge domain accompanied with relatively easy to learn image-to-image mappings to it. We have explored a special case of the edge-regularized \oursspace - specifically driving \oursspace to be a domain of edge-map-like images. In this implementation, we have shown significant advantages of our proposed approach for the important limited-source-labels tasks such as FUDA and UDG, as well as for a proposed task of generalization between cross-domain benchmarks to potentially unseen domains and classes.
We observed a significant improvement over previous unsupervised and partially supervised methods for these tasks. Future work may also include exploration of the edge-like transforms used here as potentially useful augmentations for contrastive SSL in general.

\textit{Limitations} of the current paper include: 
%
\textbf{\blue{(i)}} intentional focusing only on edge-like bridge domains, which is one of the simplest 
\oursspace one may construct. Naturally this bares limitations, e.g., lowering the relative importance of representing non-edge related features such as color. Thus, exploring other non-edge bridge domains is an important topic for future work; 
%
\textbf{\blue{(ii)}} our current approach is built on top of a very useful, yet a single SSL method, namely, MoCo \cite{chen2020improved}. A direct extension could be employing our approach on top of a vision transformer backbone using the SSL method of \cite{caron2021emerging}, or more broadly, making it applicable to any SSL method;
%
\textbf{\blue{(iii)}} finally, our approach being completely unsupervised in pre-training, lacks control over which semantic classes are formed in the learned representation space, which is a common problem shared with most current SSL techniques that might lead to missing the classes that are under-represented in terms of different instance count in the unlabeled data. Addressing this in a follow-up work may include adding such control via some form of zero-shot or few-shot priming, or by training with coarse labels \cite{Bukchin2021Fine}.

\paragraph{Acknowledgments}
This material is based upon work supported by the Defense Advanced Research Projects Agency (DARPA) under Contract No. FA8750-19-C-1001. Any opinions, findings and conclusions or recommendations expressed in this material are those of the author(s) and do not necessarily reflect the views of DARPA. Raja Giryes was supported by ERC-StG grant no. 757497 (SPADE). Dina Katabi was supported by funding from the MIT-IBM Watson AI lab.

{\small
\bibliographystyle{include/template/ieee_fullname}
\bibliography{include/brad}
}

\clearpage

\appendix
{\LARGE{\textbf{Appendix}}}

\vspace{-0.2cm}
\section{Additional edge map examples}
\label{apend:edge_maps}
Additional comparison of typical examples of \oursspace variants generated by Canny~\cite{canny1986computational}, pretrained HED~\cite{Xie_2015_ICCV}, and our `learned \ours' are shown in Figures~\ref{fig:edge_maps_real} and~\ref{fig:edge_maps_sketch}. The images are taken from PACS dataset~\cite{Li_2017_ICCV}. Figure~\ref{fig:edge_maps_real} presents images from the domains Real and Art, while Figure~\ref{fig:edge_maps_sketch} presents images from Sketch and Cartoon. As can be seen, both `learned \ours' and `HED \ours' variants discard the background noise, but unlike HED, `learned \ours' learns to retain semantic details of shape and texture. 
These retained details are intuitively highly useful for making the representations, learned using the `learned \ours' as the bridge domain $\Omega$, effective for the downstream tasks such as UDG or FUDA.
%

\vspace{-0.2cm}
\section{\oursspace loss \texorpdfstring{$\mathcal{L}_\Omega$}{} without pretrained HED}
\label{apend:wo_pre_hed}
When training \oursspace domain mappings $\Psi_n$ we utilize the \oursspace loss $\mathcal{L}_\Omega$ (Eq.~(\red{4}) in the main paper, repeated in \cref{eq:loss_l2hed} below for convenience) for distilling from an edge-mapping $\mathcal{E}$ forcing the \oursspace bridge domain $\Omega$ images to be similar to edge maps.
\begin{equation}
    \mathcal{L}_{\Omega}(I_n) = ||\Psi_n(I_n^{a1}) - \mathcal{E}(I_n^{a1})||_2^2
    \label{eq:loss_l2hed}
\end{equation}

In the main variant of our approach the $\mathcal{E}$ is a HED \cite{Xie_2015_ICCV} model pretrained on the BSDS500 dataset~\cite{amfm_pami2011} and the $\Psi_n$ models are initialized with the same pretrained model. To avoid this use of BSDS500 as additional data, we tested alternative loss functions based on Canny~\cite{canny1986computational} instead of HED~\cite{Xie_2015_ICCV}, while randomly initializing $\Psi_n$.
Since the edges in Canny edge-maps are only one pixel wide we apply a Gaussian blurring before comparing to the current $\Psi_n$ output. In our implementation we used a blur kernel of size $5$ with $\sigma=0.15$. We test both L1 and L2 norms, however, when using L1-norm we first stretch the $\Psi_n$ output to the range [0,1] for stability. \cref{eq:loss_l2canny} and~\cref{eq:loss_l1canny} present both these variants of the Canny-based $\mathcal{L}_{\Omega}$ loss functions. 

\begin{equation}
    \mathcal{L}_{\Omega}(I_n) = ||\Psi_n(I_n^{a1}) - Blur(\mathcal{E}_{Canny}(I_n^{a1}))||_2^2
    \label{eq:loss_l2canny}
\end{equation}

\begin{figure}[h]
\vspace{0.3cm}
     \centering
     \begin{subfigure}[b]{0.2\textwidth}
         \centering
         \includegraphics[width=\textwidth]{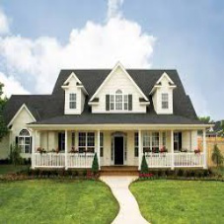}
         \caption{Original image}
     \end{subfigure}
     \begin{subfigure}[b]{0.2\textwidth}
         \centering
         \includegraphics[width=\textwidth]{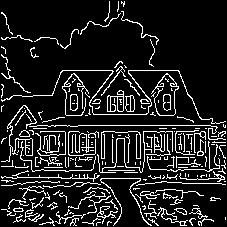}
         \caption{Canny}
         \label{subfig:Canny}
     \end{subfigure}
     \begin{subfigure}[b]{0.2\textwidth}
         \centering
         \includegraphics[width=\textwidth]{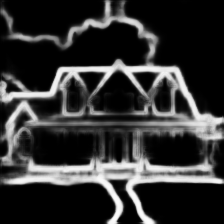}
         \caption{Hed}
          \label{subfig:Hed}
     \end{subfigure}
          \begin{subfigure}[b]{0.2\textwidth}
         \centering
         \includegraphics[width=\textwidth]{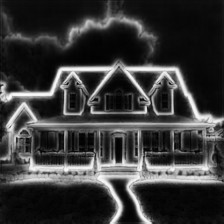}
         \caption{L2 Hed-based $L_\Omega$ (Eq. \ref{eq:loss_l2hed})}
         \label{subfig:L2Hed}
     \end{subfigure}
          \begin{subfigure}[b]{0.2\textwidth}
         \centering
         \includegraphics[width=\textwidth]{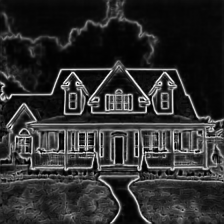}
         \caption{L2 Canny-based $L_\Omega$ (Eq. \ref{eq:loss_l2canny})}
         \label{subfig:L2Canny}
     \end{subfigure}     
          \begin{subfigure}[b]{0.2\textwidth}
         \centering
         \includegraphics[width=\textwidth]{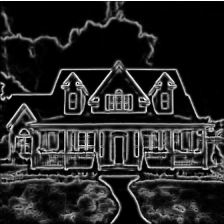}
         \caption{L1 Canny-based $L_\Omega$ (Eq. \ref{eq:loss_l1canny})}
         \label{subfig:L1Canny}
     \end{subfigure}     

    \caption{Output images of \oursspace mapping functions $\Psi_n$ trained using different loss functions $L_\Omega$ also compared to Canny~\cite{canny1986computational} and HED~\cite{Xie_2015_ICCV} edge maps. Canny produces noisy images with many irrelevant edges, while HED mostly outlines the object and discards important internal texture. All learned $\Psi_n$ retain semantic details of shape and texture better than HED while discarding most of the noise. Please zoom.}
    \label{fig:edge_images}
    \vspace{-0.3cm}
\end{figure}

\vspace{-0.6cm}

\begin{equation}
    \mathcal{L}_{\Omega}(I_n) = ||\mathcal{S}(\Psi_n(I_n^{a1})) - Blur(\mathcal{E}_{Canny}(I_n^{a1}))||_1
    \label{eq:loss_l1canny}
\end{equation}
where $\mathcal{S}$ is a pixel-wise stretch function to the range [0,1].
\cref{tab:ablation_loss} presents the average FUDA accuracy results on DomainNet using the above loss functions. As can be seen the differences in performance are quite small. \cref{fig:edge_images} presents examples of \oursspace images using the different losses. As can be clearly seen, all the learned \oursspace variants (\cref{fig:edge_images}d-f) retain semantic details of shape and texture better than the fixed a-priori \oursspace variants (\cref{fig:edge_images}b-c).

\begin{table}[t]
\small
    \centering
    \begin{tabular}{l|cc}
        \toprule
        Loss  & 1-shot & 3-shots \\
        \midrule
         L2 Hed (equation \ref{eq:loss_l2hed})  & \textbf{48.64}  & 58.31\\
         L2 Canny (equation \ref{eq:loss_l2canny}) & 47.40  & \textbf{59.30}\\
         L1 Canny (equation \ref{eq:loss_l1canny}) & \textcolor{blue}{47.77}  & \textcolor{blue}{58.66}\\
        \bottomrule
    \end{tabular}
    \caption{FUDA accuracy (\%) results on DomainNet using different $L_\Omega$ \oursspace losses. \textbf{bold} = best, \textcolor{blue}{blue} = second best.}
    
    \label{tab:ablation_loss}
    \tabvspace
\end{table}



\newcommand*\figsize{0.15\textwidth}

\begin{figure*}[ht]
    \centering
     \begin{subfigure}[b]{\figsize}
         \centering
         \caption{Original}
         \includegraphics[width=\textwidth]{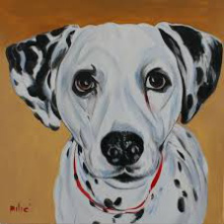}
    \end{subfigure}
     \centering
     \begin{subfigure}[b]{\figsize}
         \centering
         \caption{Canny}
         \includegraphics[width=\textwidth]{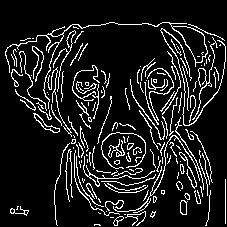}
    \end{subfigure}
     \centering
     \begin{subfigure}[b]{\figsize}
         \centering
         \caption{HED}
         \includegraphics[width=\textwidth]{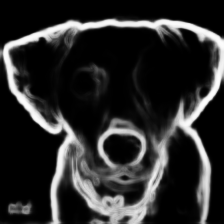}
    \end{subfigure}
     \centering
     \begin{subfigure}[b]{\figsize}
         \centering
         \caption{Learned \ours}
         \includegraphics[width=\textwidth]{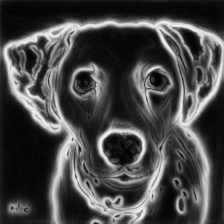}
    \end{subfigure}

           \centering
     \begin{subfigure}[b]{\figsize}
         \centering
         \includegraphics[width=\textwidth]{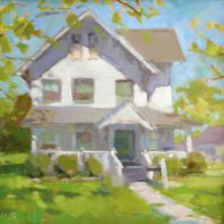}
    \end{subfigure}
     \centering
     \begin{subfigure}[b]{\figsize}
         \centering
         \includegraphics[width=\textwidth]{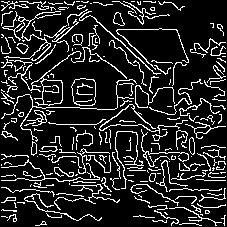}
    \end{subfigure}
     \centering
     \begin{subfigure}[b]{\figsize}
         \centering
         \includegraphics[width=\textwidth]{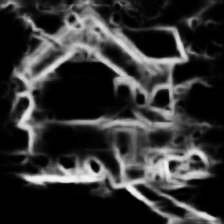}
    \end{subfigure}
     \centering
     \begin{subfigure}[b]{\figsize}
         \centering
         \includegraphics[width=\textwidth]{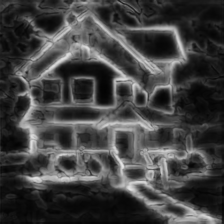}
    \end{subfigure}
    
    \centering
     \begin{subfigure}[b]{\figsize}
         \centering
         \includegraphics[width=\textwidth]{figures/edge_images/giraffe_art_orig.png}
    \end{subfigure}
     \centering
     \begin{subfigure}[b]{\figsize}
         \centering
         \includegraphics[width=\textwidth]{figures/edge_images/giraffe_art_canny.png}
    \end{subfigure}
     \centering
     \begin{subfigure}[b]{\figsize}
         \centering
         \includegraphics[width=\textwidth]{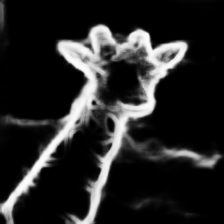}
    \end{subfigure}
     \centering
     \begin{subfigure}[b]{\figsize}
         \centering
         \includegraphics[width=\textwidth]{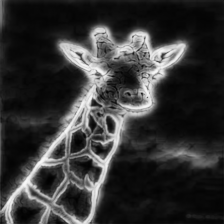}
    \end{subfigure}

        \centering
     \begin{subfigure}[b]{\figsize}
         \centering
         \includegraphics[width=\textwidth]{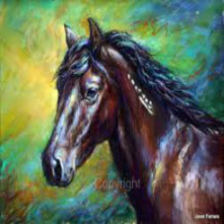}
    \end{subfigure}
     \centering
     \begin{subfigure}[b]{\figsize}
         \centering
         \includegraphics[width=\textwidth]{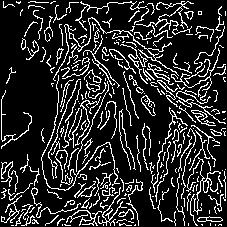}
    \end{subfigure}
     \centering
     \begin{subfigure}[b]{\figsize}
         \centering
         \includegraphics[width=\textwidth]{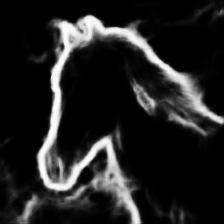}
    \end{subfigure}
     \centering
     \begin{subfigure}[b]{\figsize}
         \centering
         \includegraphics[width=\textwidth]{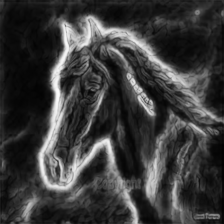}
    \end{subfigure}

            \centering
     \begin{subfigure}[b]{\figsize}
         \centering
         \includegraphics[width=\textwidth]{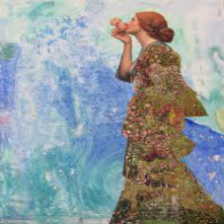}
    \end{subfigure}
     \centering
     \begin{subfigure}[b]{\figsize}
         \centering
         \includegraphics[width=\textwidth]{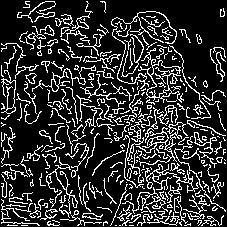}
    \end{subfigure}
     \centering
     \begin{subfigure}[b]{\figsize}
         \centering
         \includegraphics[width=\textwidth]{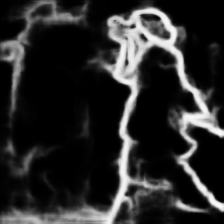}
    \end{subfigure}
     \centering
     \begin{subfigure}[b]{\figsize}
         \centering
         \includegraphics[width=\textwidth]{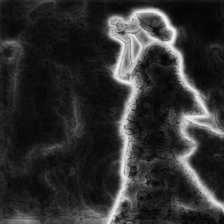}
    \end{subfigure}

            \centering
     \begin{subfigure}[b]{\figsize}
         \centering
         \includegraphics[width=\textwidth]{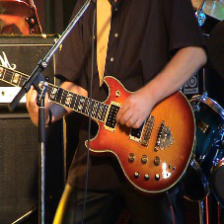}
    \end{subfigure}
     \centering
     \begin{subfigure}[b]{\figsize}
         \centering
         \includegraphics[width=\textwidth]{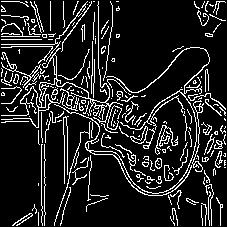}
    \end{subfigure}
     \centering
     \begin{subfigure}[b]{\figsize}
         \centering
         \includegraphics[width=\textwidth]{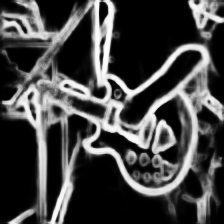}
    \end{subfigure}
     \centering
     \begin{subfigure}[b]{\figsize}
         \centering
         \includegraphics[width=\textwidth]{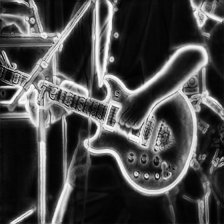}
    \end{subfigure}

   \centering
     \begin{subfigure}[b]{\figsize}
         \centering
         \includegraphics[width=\textwidth]{figures/edge_images/house_real2_orig.png}
    \end{subfigure}
     \centering
     \begin{subfigure}[b]{\figsize}
         \centering
         \includegraphics[width=\textwidth]{figures/edge_images/house_real2_canny.jpg}
    \end{subfigure}
     \centering
     \begin{subfigure}[b]{\figsize}
         \centering
         \includegraphics[width=\textwidth]{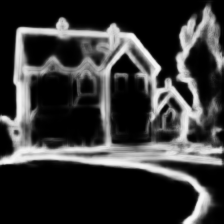}
    \end{subfigure}
     \centering
     \begin{subfigure}[b]{\figsize}
         \centering
         \includegraphics[width=\textwidth]{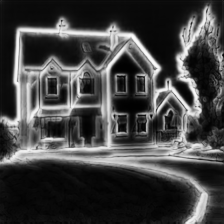}
    \end{subfigure}

               \centering
     \begin{subfigure}[b]{\figsize}
         \centering
         \includegraphics[width=\textwidth]{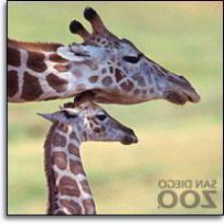}
    \end{subfigure}
     \centering
     \begin{subfigure}[b]{\figsize}
         \centering
         \includegraphics[width=\textwidth]{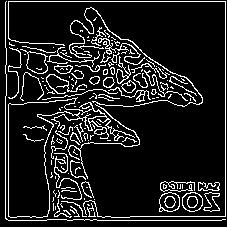}
    \end{subfigure}
     \centering
     \begin{subfigure}[b]{\figsize}
         \centering
         \includegraphics[width=\textwidth]{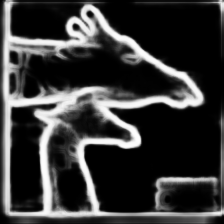}
    \end{subfigure}
     \centering
     \begin{subfigure}[b]{\figsize}
         \centering
         \includegraphics[width=\textwidth]{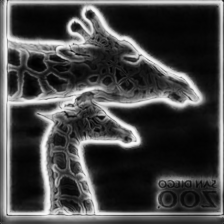}
    \end{subfigure}

    \caption{Edge images for different choices of $\Psi_n$. Images are taken from Real and Art domains of PACS dataset~\cite{Li_2017_ICCV}.}
    \label{fig:edge_maps_real}
\end{figure*}

\begin{figure*}[ht]

    \centering
     \begin{subfigure}[b]{\figsize}
         \centering
         \caption{Original}
         \includegraphics[width=\textwidth]{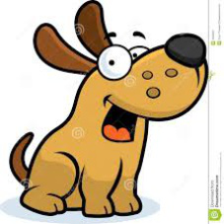}
    \end{subfigure}
     \centering
     \begin{subfigure}[b]{\figsize}
         \centering
         \caption{Canny}
         \includegraphics[width=\textwidth]{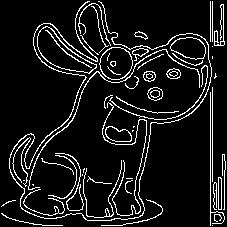}
    \end{subfigure}
     \centering
     \begin{subfigure}[b]{\figsize}
         \centering
         \caption{HED}
         \includegraphics[width=\textwidth]{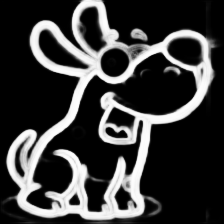}
    \end{subfigure}
     \centering
     \begin{subfigure}[b]{\figsize}
         \centering
         \caption{Learned \ours}
         \includegraphics[width=\textwidth]{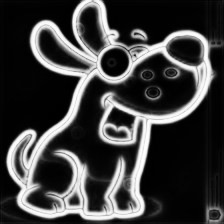}
    \end{subfigure}

        \centering
     \begin{subfigure}[b]{\figsize}
         \centering
         \includegraphics[width=\textwidth]{figures/edge_images/person_sketch_orig.png}
    \end{subfigure}
     \centering
     \begin{subfigure}[b]{\figsize}
         \centering
         \includegraphics[width=\textwidth]{figures/edge_images/person_sketch_canny.png}
    \end{subfigure}
     \centering
     \begin{subfigure}[b]{\figsize}
         \centering
         \includegraphics[width=\textwidth]{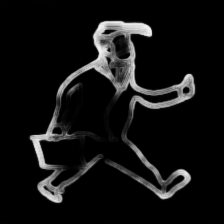}
    \end{subfigure}
     \centering
     \begin{subfigure}[b]{\figsize}
         \centering
         \includegraphics[width=\textwidth]{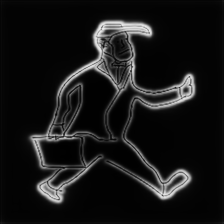}
    \end{subfigure}

        \centering
     \begin{subfigure}[b]{\figsize}
         \centering
         \includegraphics[width=\textwidth]{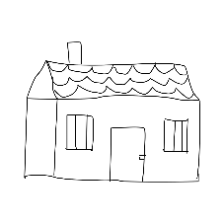}
    \end{subfigure}
     \centering
     \begin{subfigure}[b]{\figsize}
         \centering
         \includegraphics[width=\textwidth]{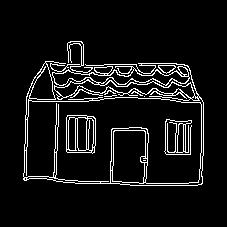}
    \end{subfigure}
     \centering
     \begin{subfigure}[b]{\figsize}
         \centering
         \includegraphics[width=\textwidth]{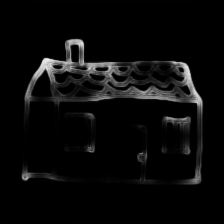}
    \end{subfigure}
     \centering
     \begin{subfigure}[b]{\figsize}
         \centering
         \includegraphics[width=\textwidth]{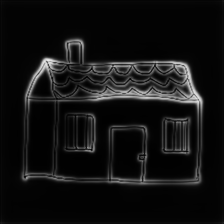}
    \end{subfigure}

        \centering
     \begin{subfigure}[b]{\figsize}
         \centering
         \includegraphics[width=\textwidth]{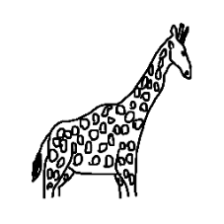}
    \end{subfigure}
     \centering
     \begin{subfigure}[b]{\figsize}
         \centering
         \includegraphics[width=\textwidth]{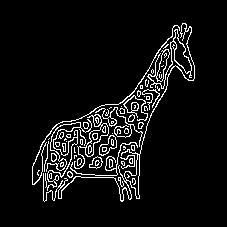}
    \end{subfigure}
     \centering
     \begin{subfigure}[b]{\figsize}
         \centering
         \includegraphics[width=\textwidth]{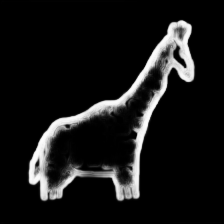}
    \end{subfigure}
     \centering
     \begin{subfigure}[b]{\figsize}
         \centering
         \includegraphics[width=\textwidth]{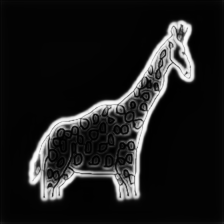}
    \end{subfigure}

        \centering
     \begin{subfigure}[b]{\figsize}
         \centering
         \includegraphics[width=\textwidth]{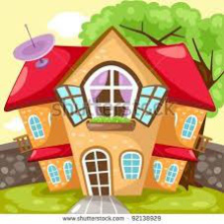}
    \end{subfigure}
     \centering
     \begin{subfigure}[b]{\figsize}
         \centering
         \includegraphics[width=\textwidth]{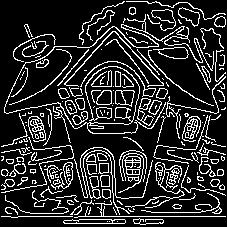}
    \end{subfigure}
     \centering
     \begin{subfigure}[b]{\figsize}
         \centering
         \includegraphics[width=\textwidth]{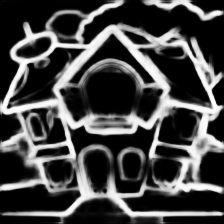}
    \end{subfigure}
     \centering
     \begin{subfigure}[b]{\figsize}
         \centering
         \includegraphics[width=\textwidth]{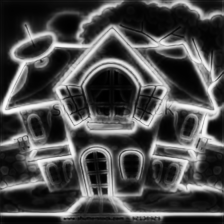}
    \end{subfigure}
    
           \centering
     \begin{subfigure}[b]{\figsize}
         \centering
         \includegraphics[width=\textwidth]{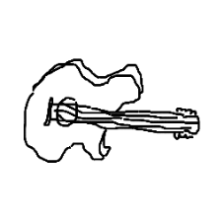}
    \end{subfigure}
     \centering
     \begin{subfigure}[b]{\figsize}
         \centering
         \includegraphics[width=\textwidth]{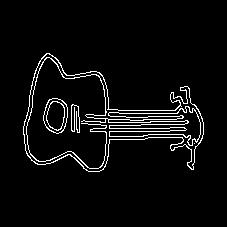}
    \end{subfigure}
     \centering
     \begin{subfigure}[b]{\figsize}
         \centering
         \includegraphics[width=\textwidth]{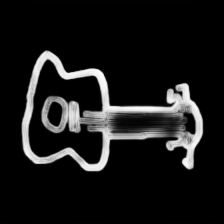}
    \end{subfigure}
     \centering
     \begin{subfigure}[b]{\figsize}
         \centering
         \includegraphics[width=\textwidth]{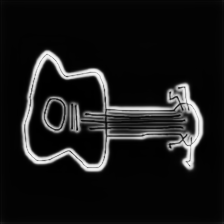}
    \end{subfigure}

          \centering
     \begin{subfigure}[b]{\figsize}
         \centering
         \includegraphics[width=\textwidth]{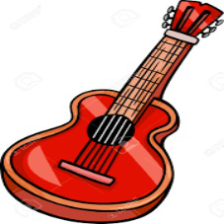}
    \end{subfigure}
     \centering
     \begin{subfigure}[b]{\figsize}
         \centering
         \includegraphics[width=\textwidth]{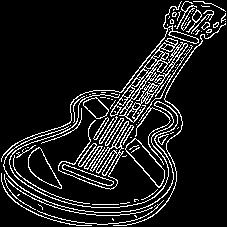}
    \end{subfigure}
     \centering
     \begin{subfigure}[b]{\figsize}
         \centering
         \includegraphics[width=\textwidth]{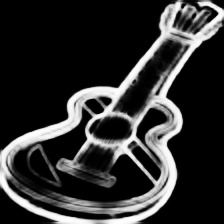}
    \end{subfigure}
     \centering
     \begin{subfigure}[b]{\figsize}
         \centering
         \includegraphics[width=\textwidth]{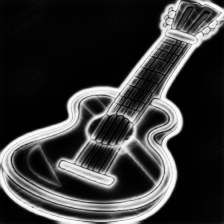}
    \end{subfigure}

    \caption{Edge images for different choices of $\Psi_n$. Images are taken from Cartoon and Sketch domains of PACS dataset~\cite{Li_2017_ICCV}.}
    \label{fig:edge_maps_sketch}
\end{figure*}

\section{Demo}
\label{apend:demo}
\begin{figure}
    \centering
    \includegraphics[width=\linewidth]{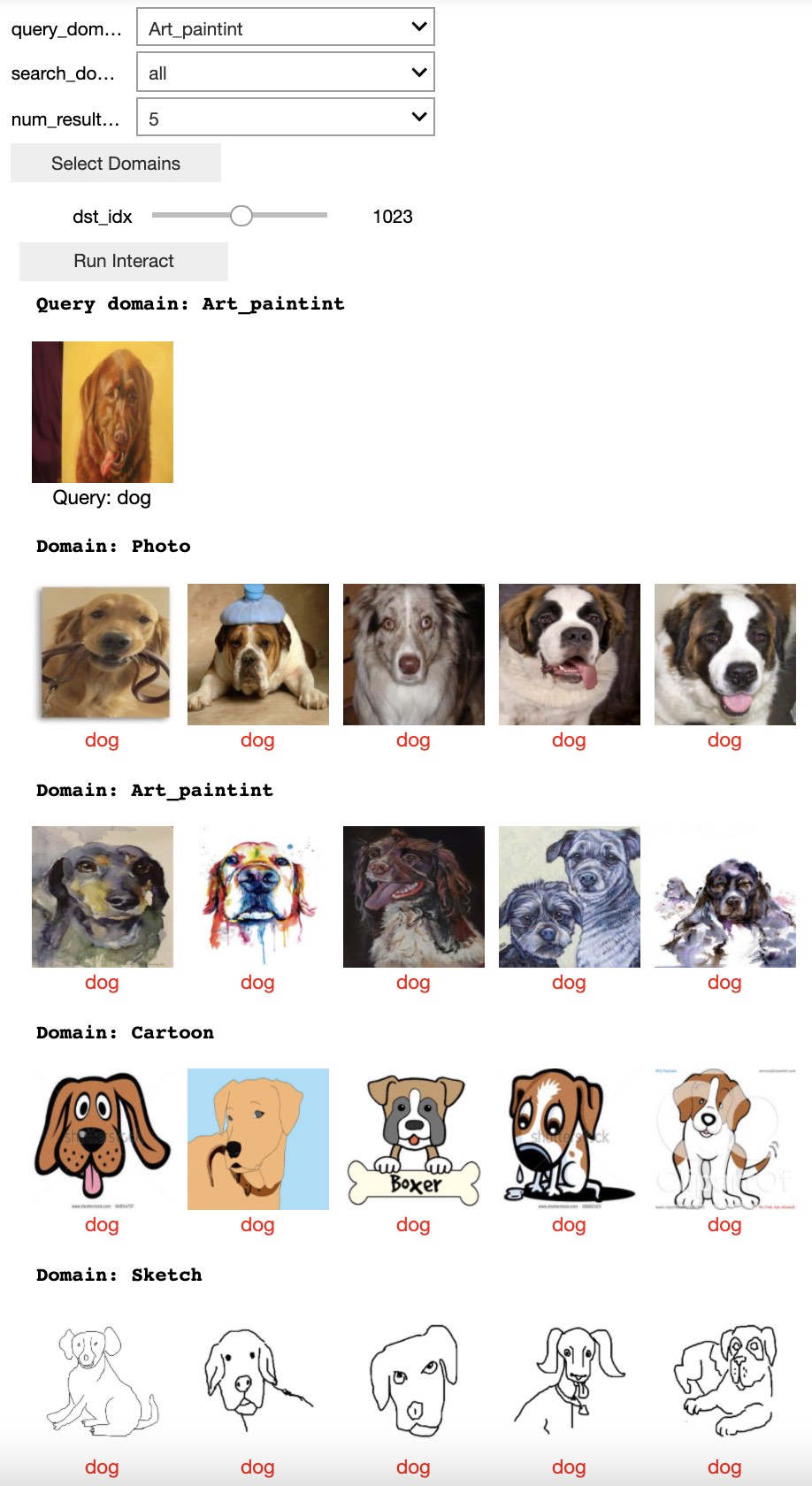}
    \caption{\democap{Dog}}
    \label{fig:demo1}
\end{figure}

\begin{figure}
    \centering
    \includegraphics[width=\linewidth]{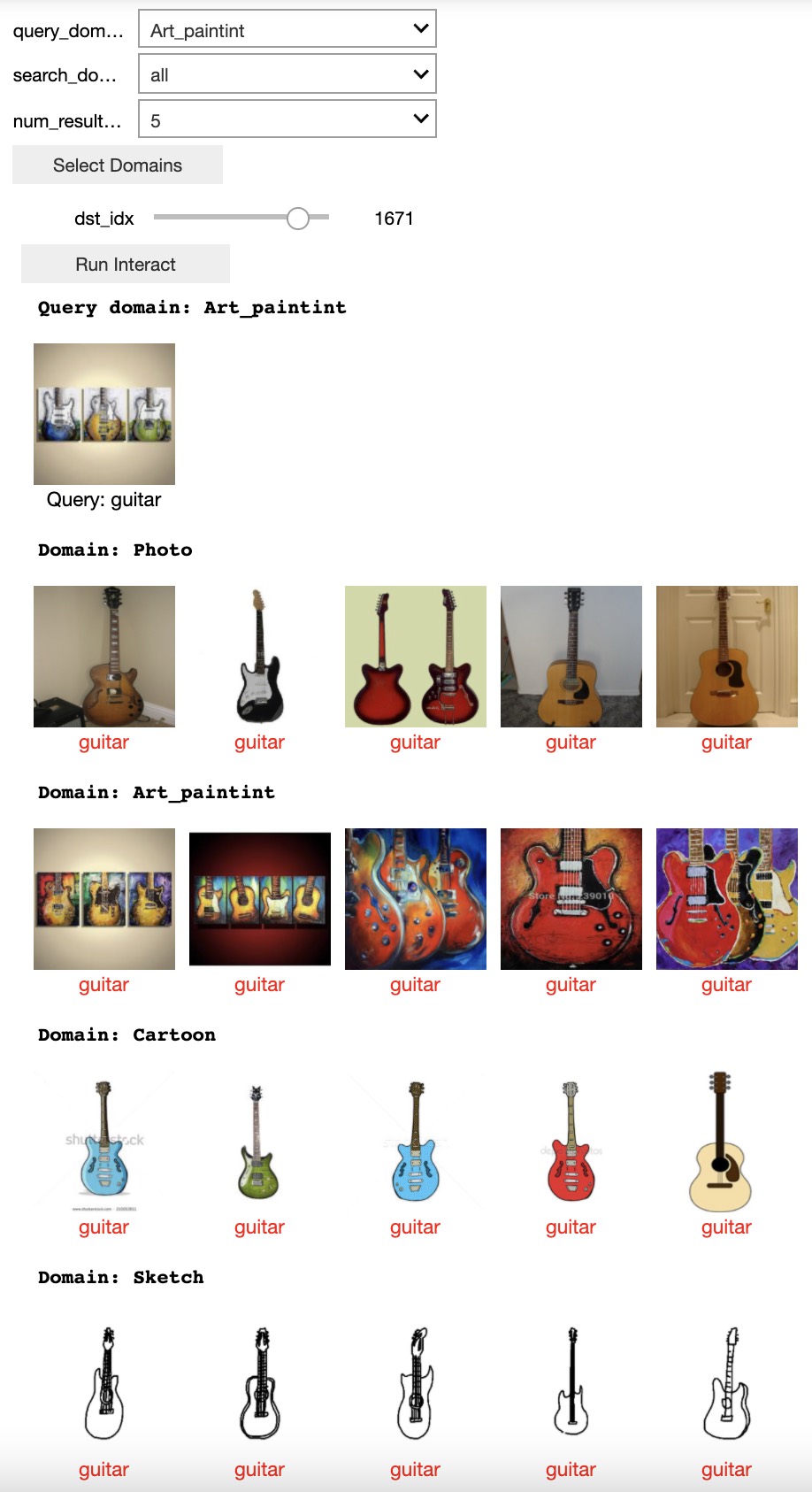}
    \caption{\democap{Guitar}}
    \label{fig:demo2}
\end{figure}

\begin{figure}
    \centering
    \includegraphics[width=\linewidth]{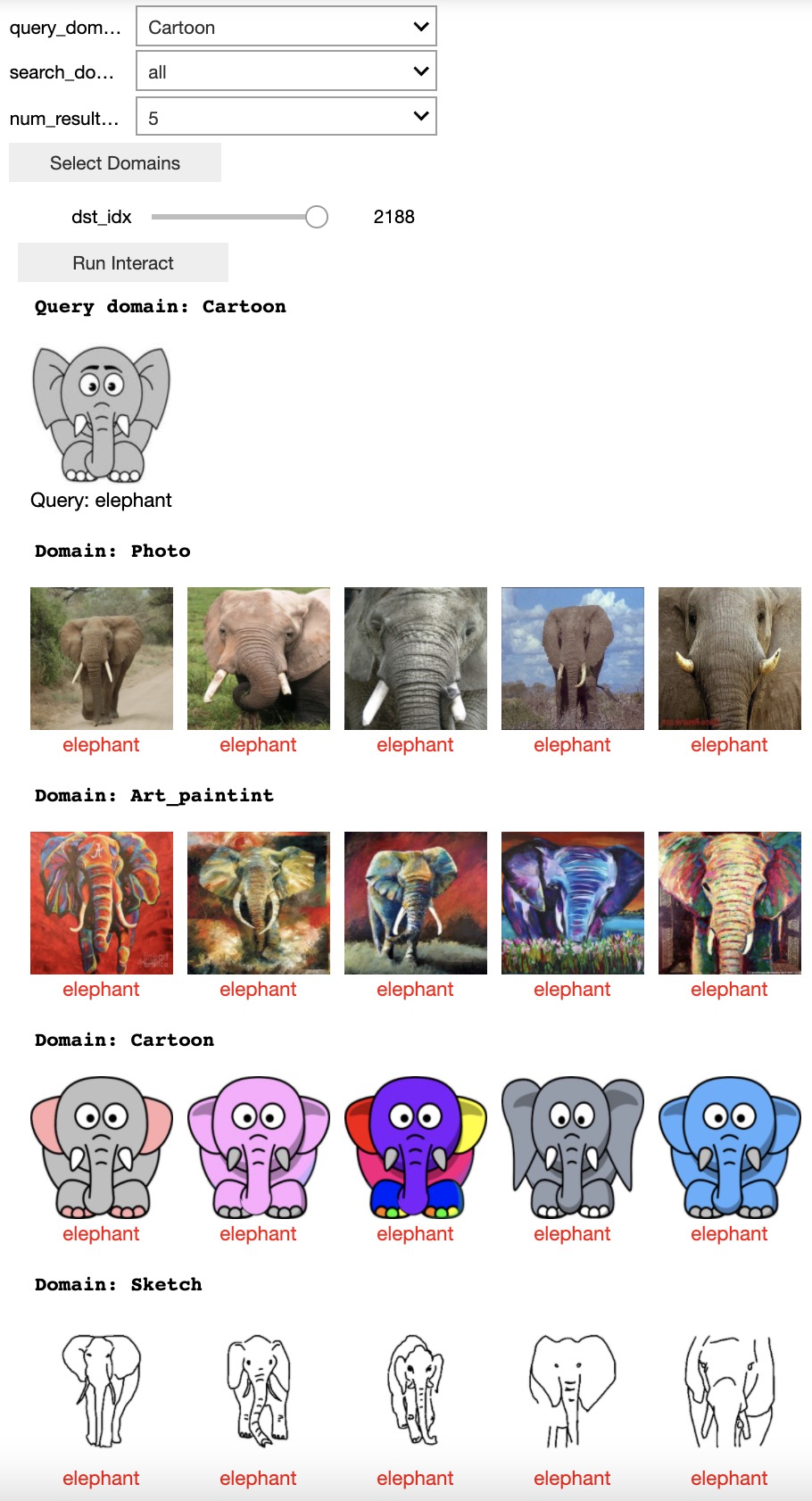}
    \caption{\democap{Elephant}}
    \label{fig:demo3}
\end{figure}

\begin{figure}
    \centering
    \includegraphics[width=\linewidth]{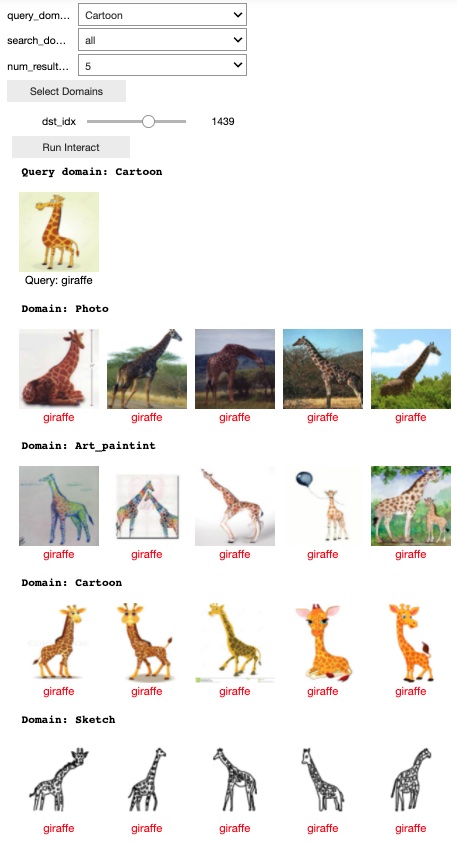}
    \caption{\democap{Giraffe}}
    \label{fig:demo4}
\end{figure}

\begin{figure}
    \centering
    \includegraphics[width=\linewidth]{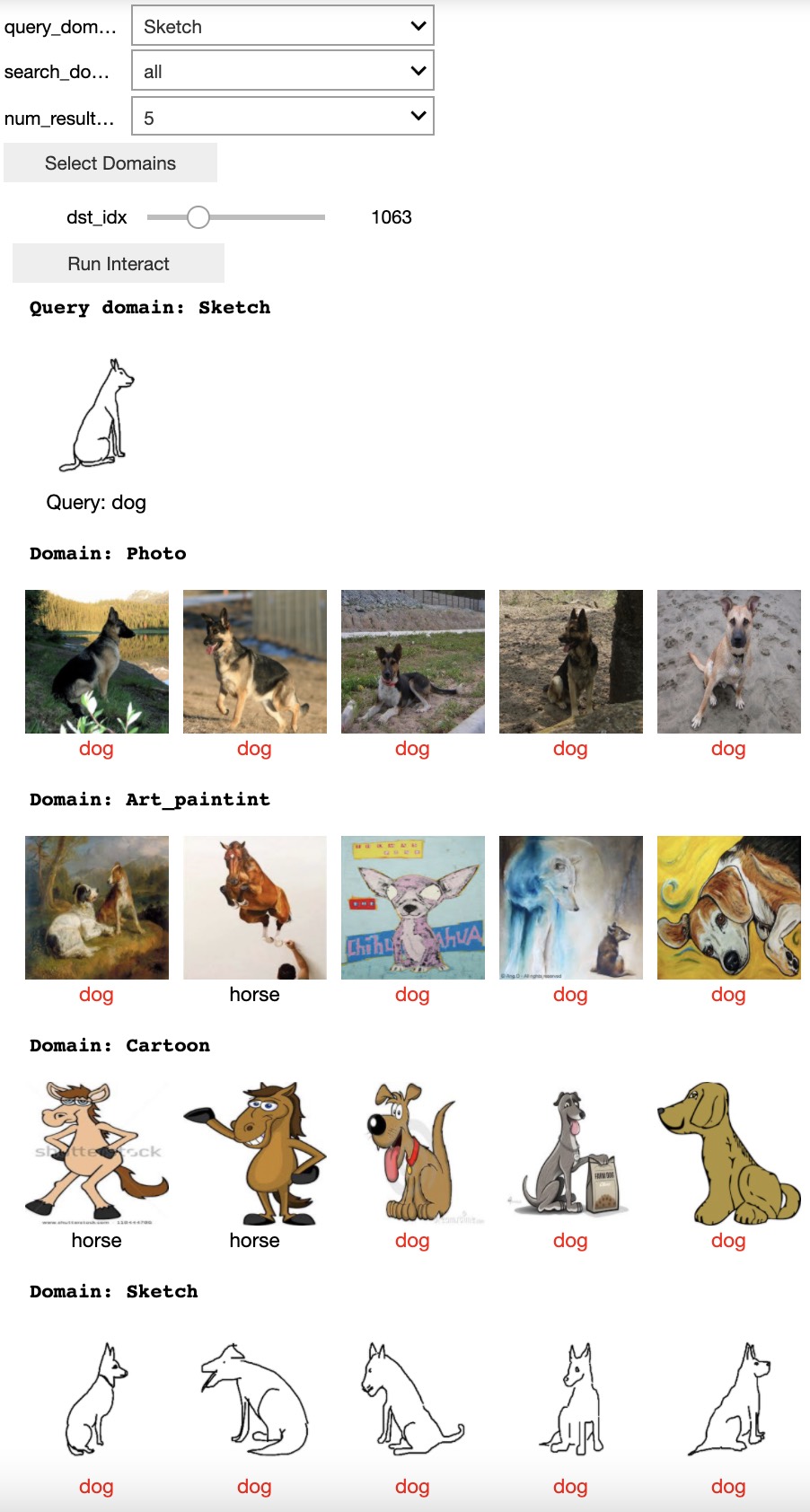}
    \caption{\democap{Dog}}
    \label{fig:demo5}
\end{figure}

\begin{figure}
    \centering
    \includegraphics[width=\linewidth]{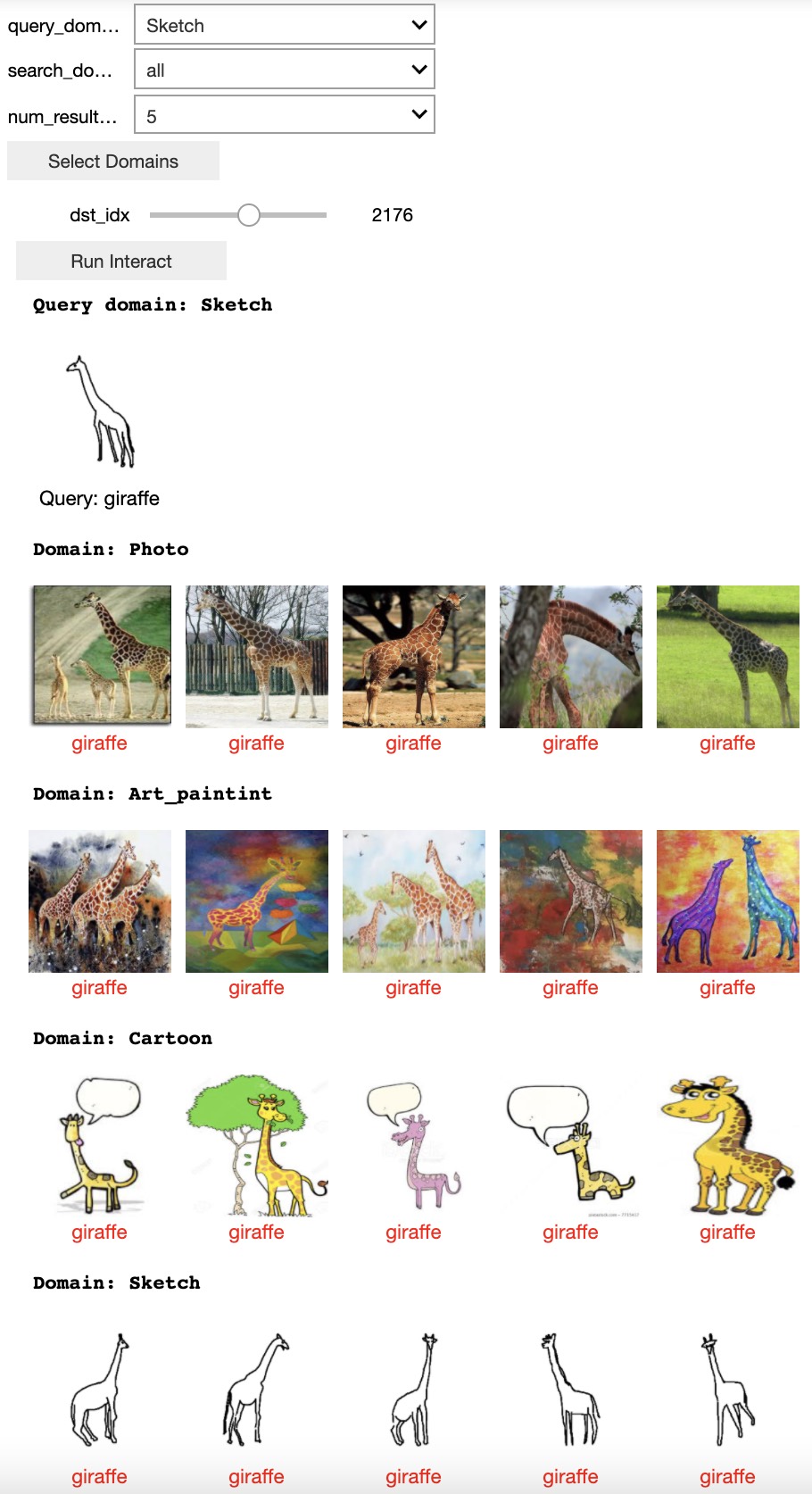}
    \caption{\democap{Giraffe}}
    \label{fig:demo6}
\end{figure}

\begin{figure}
    \centering
    \includegraphics[width=\linewidth]{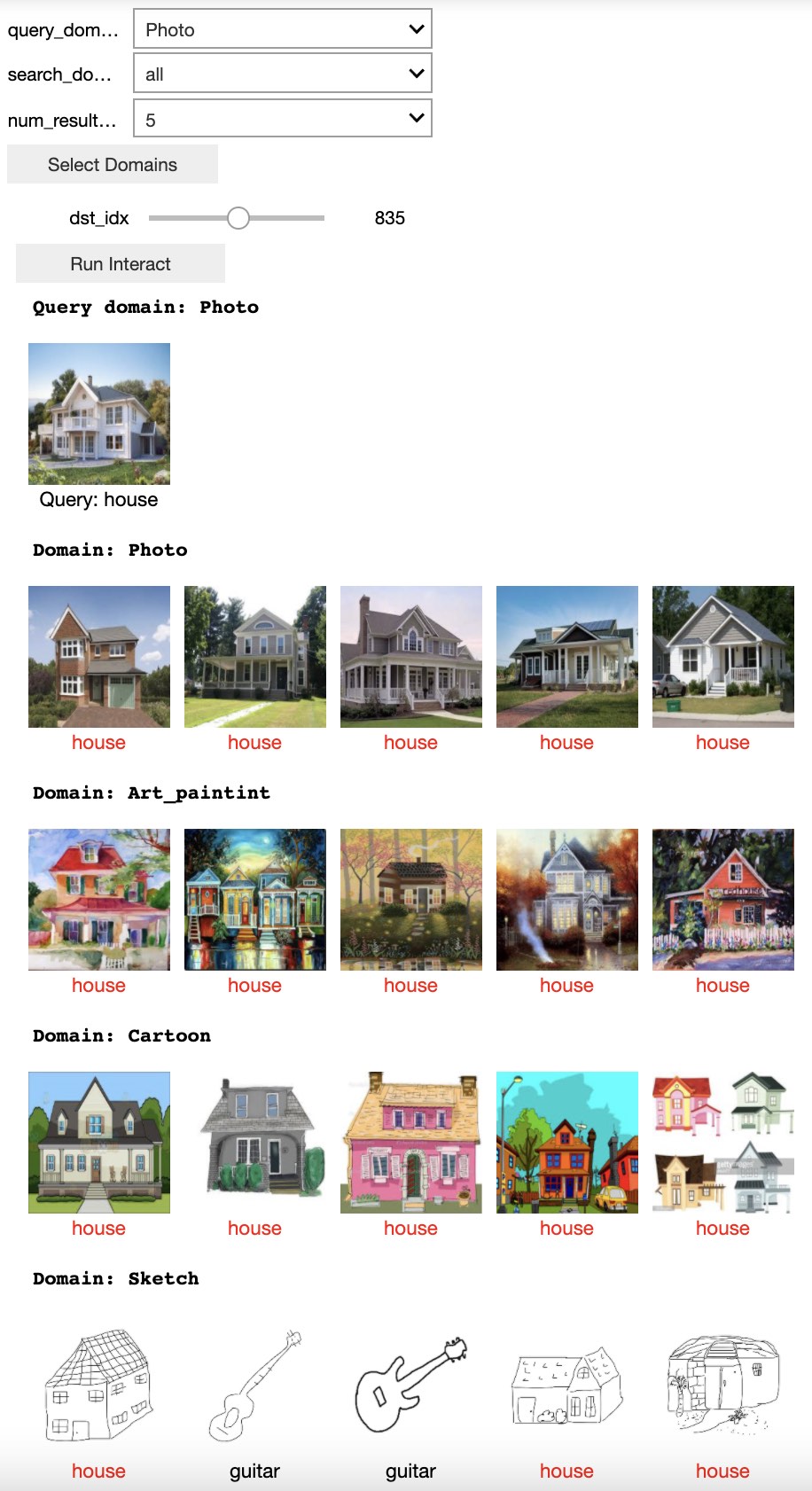}
    \caption{\democap{House}}
    \label{fig:demo7}
\end{figure}

\begin{figure}
    \centering
    \includegraphics[width=\linewidth]{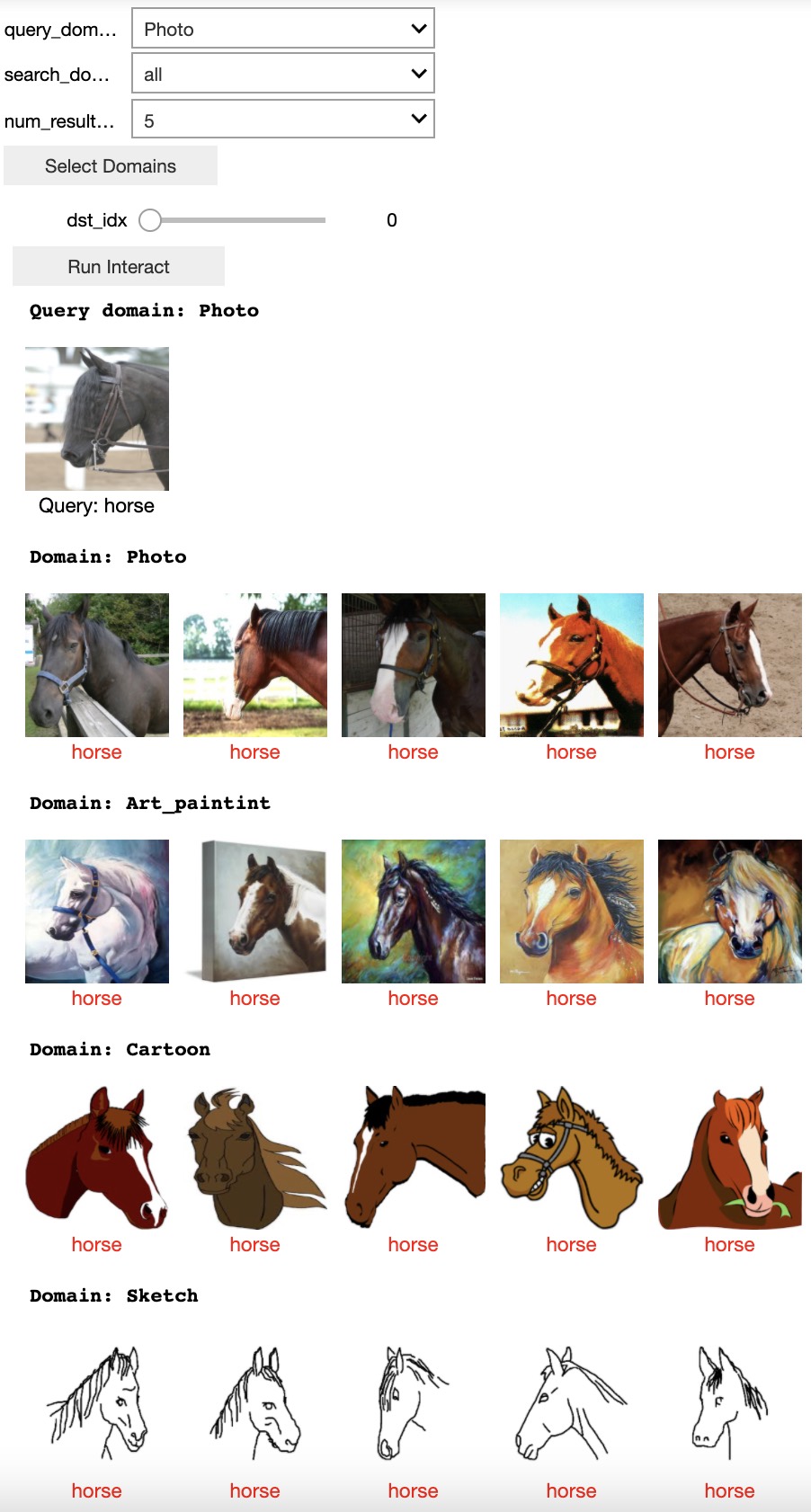}
    \caption{\democap{Horse}}
    \label{fig:demo8}
\end{figure}

\begin{figure}
    \centering
    \includegraphics[width=\linewidth]{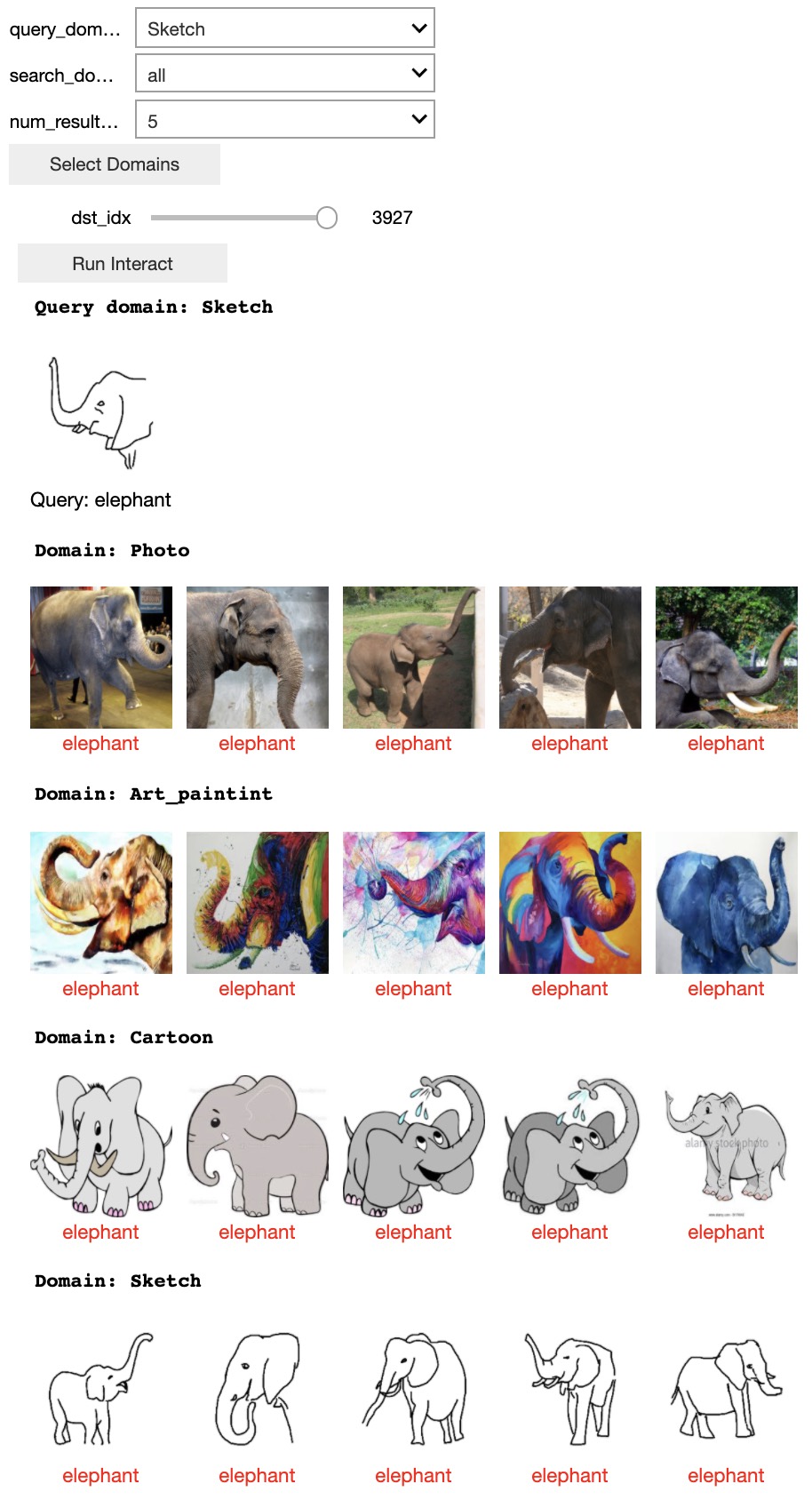}
    \caption{\democap{Elephant}}
    \label{fig:demo9}
\end{figure}

\begin{figure}
    \centering
    \includegraphics[width=\linewidth]{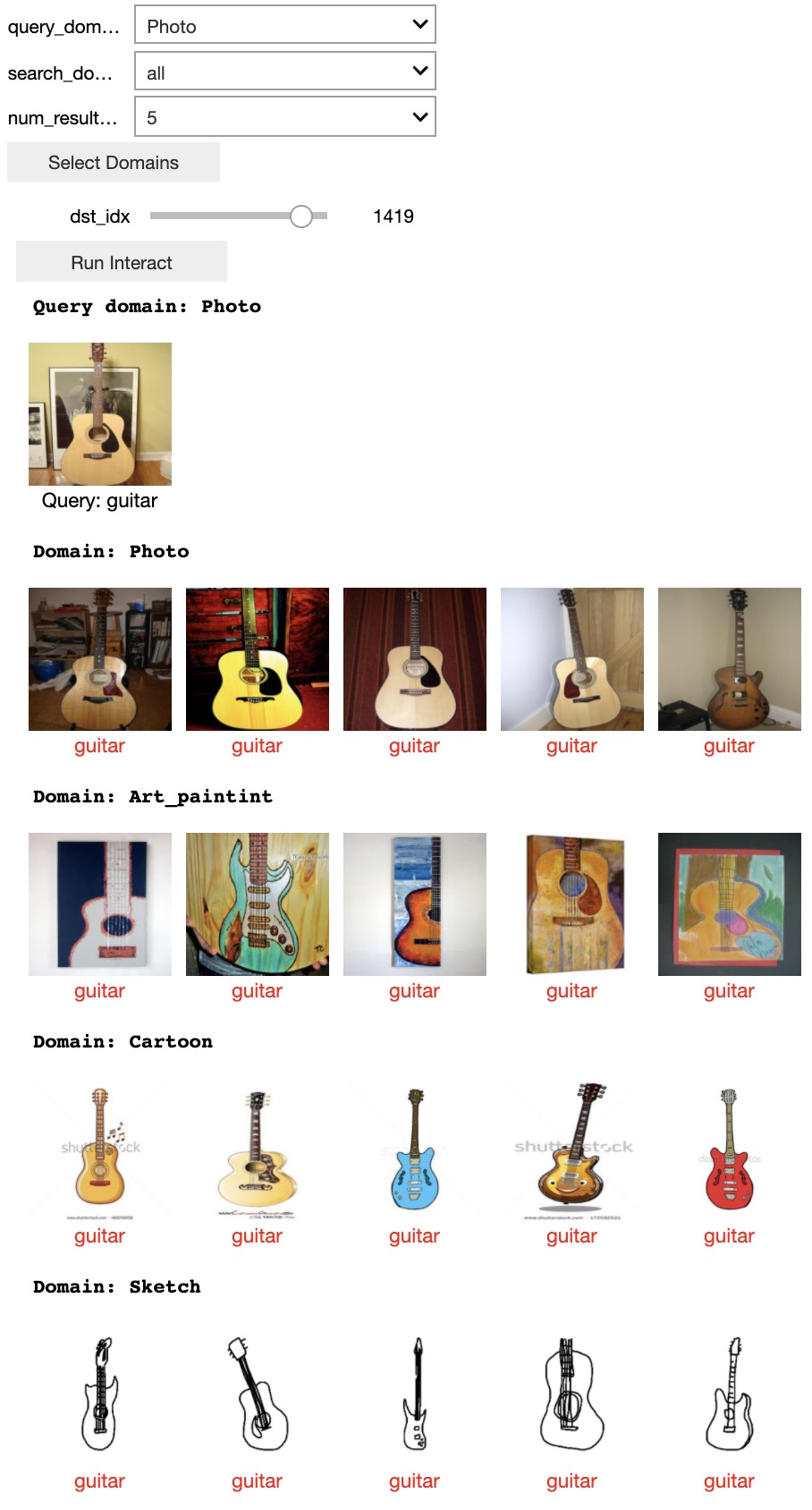}
    \caption{\democap{Guitar}}
    \label{fig:demo10}
\end{figure}

\begin{figure}
    \centering
    \includegraphics[width=\linewidth]{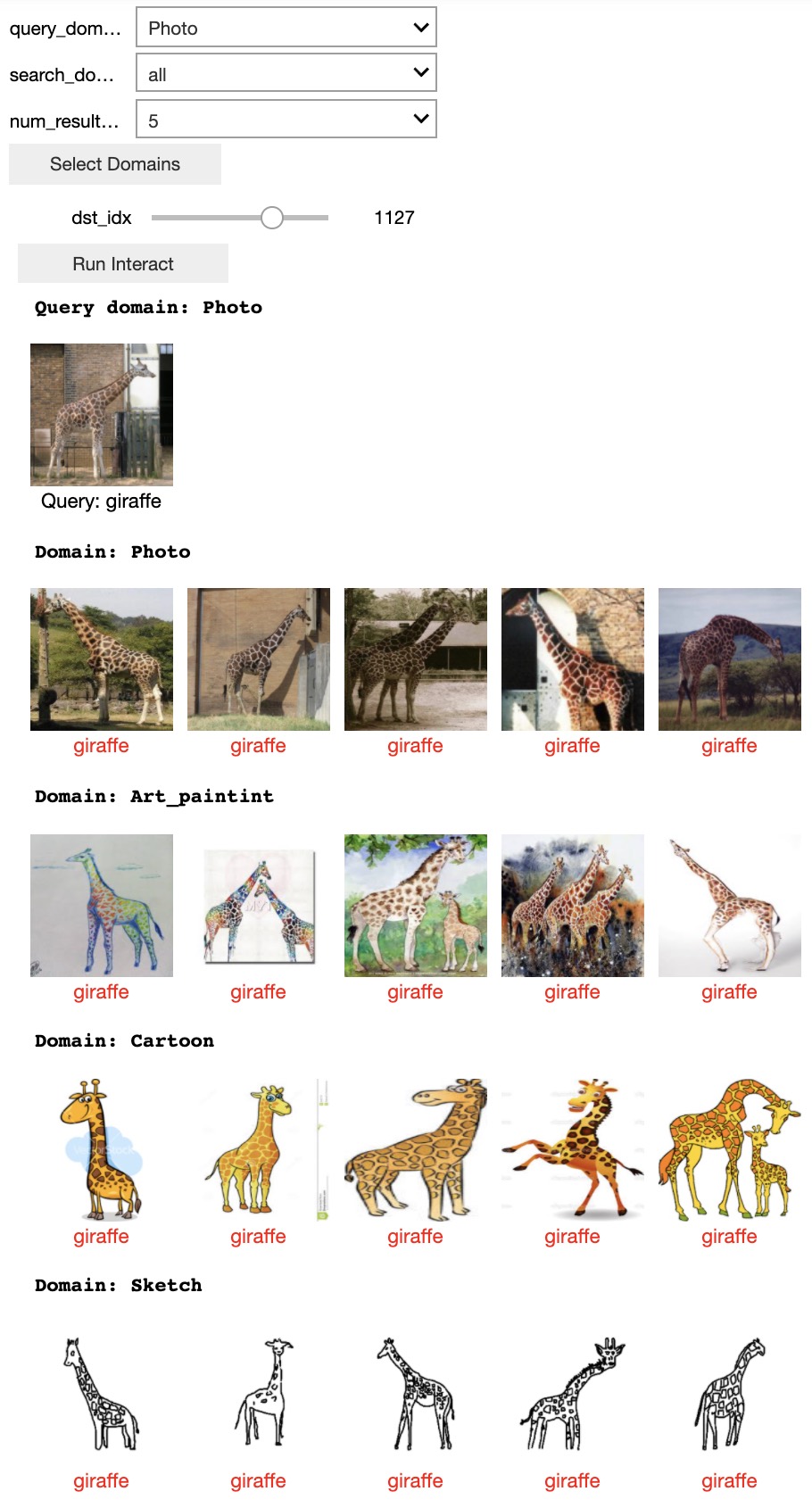}
    \caption{\democap{Giraffe}}
    \label{fig:demo11}
\end{figure}

\begin{figure}
    \centering
    \includegraphics[width=\linewidth]{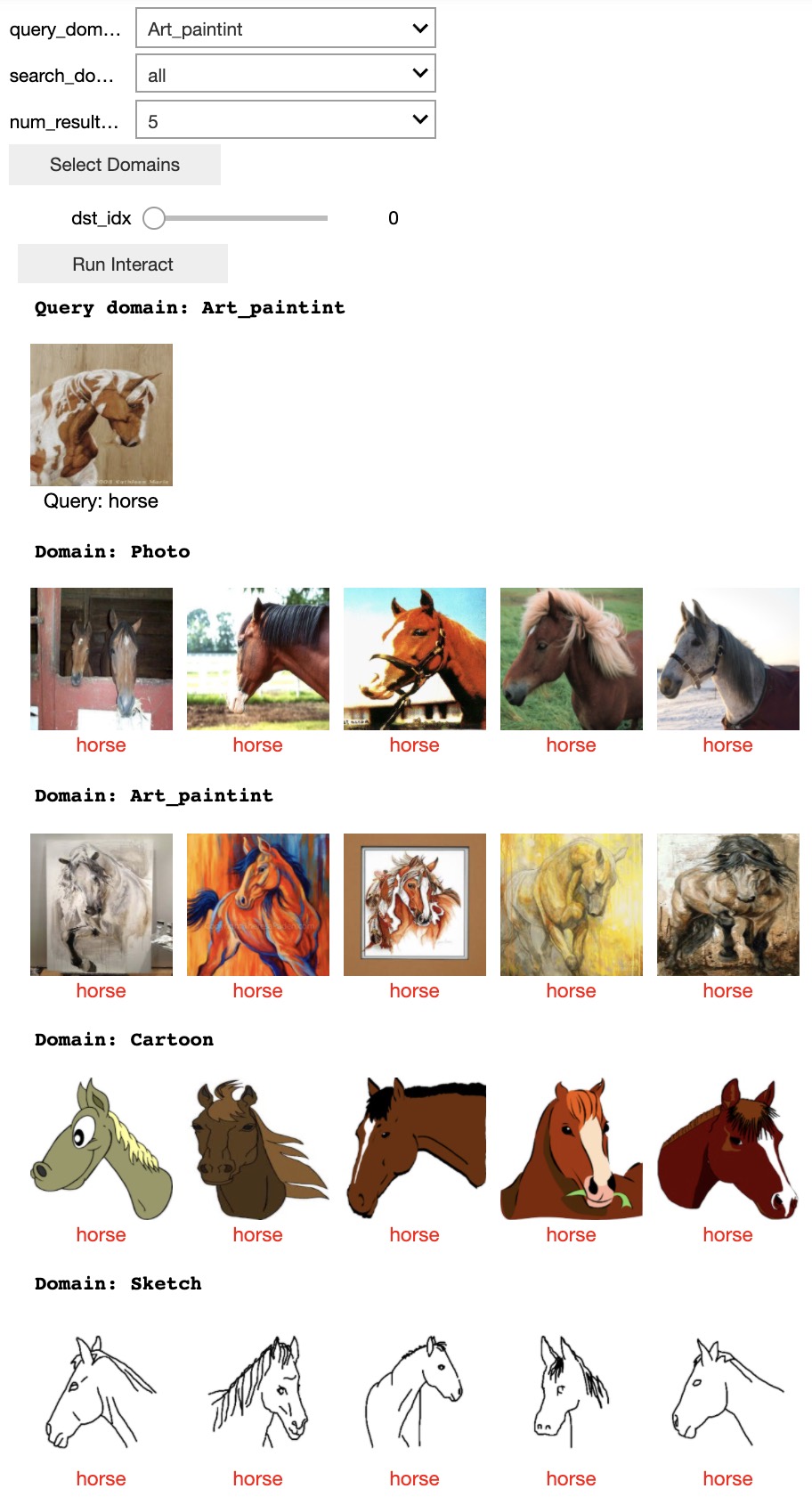}
    \caption{\democap{Horse}}
    \label{fig:demo12}
\end{figure}

\begin{figure}
    \centering
    \includegraphics[width=\linewidth]{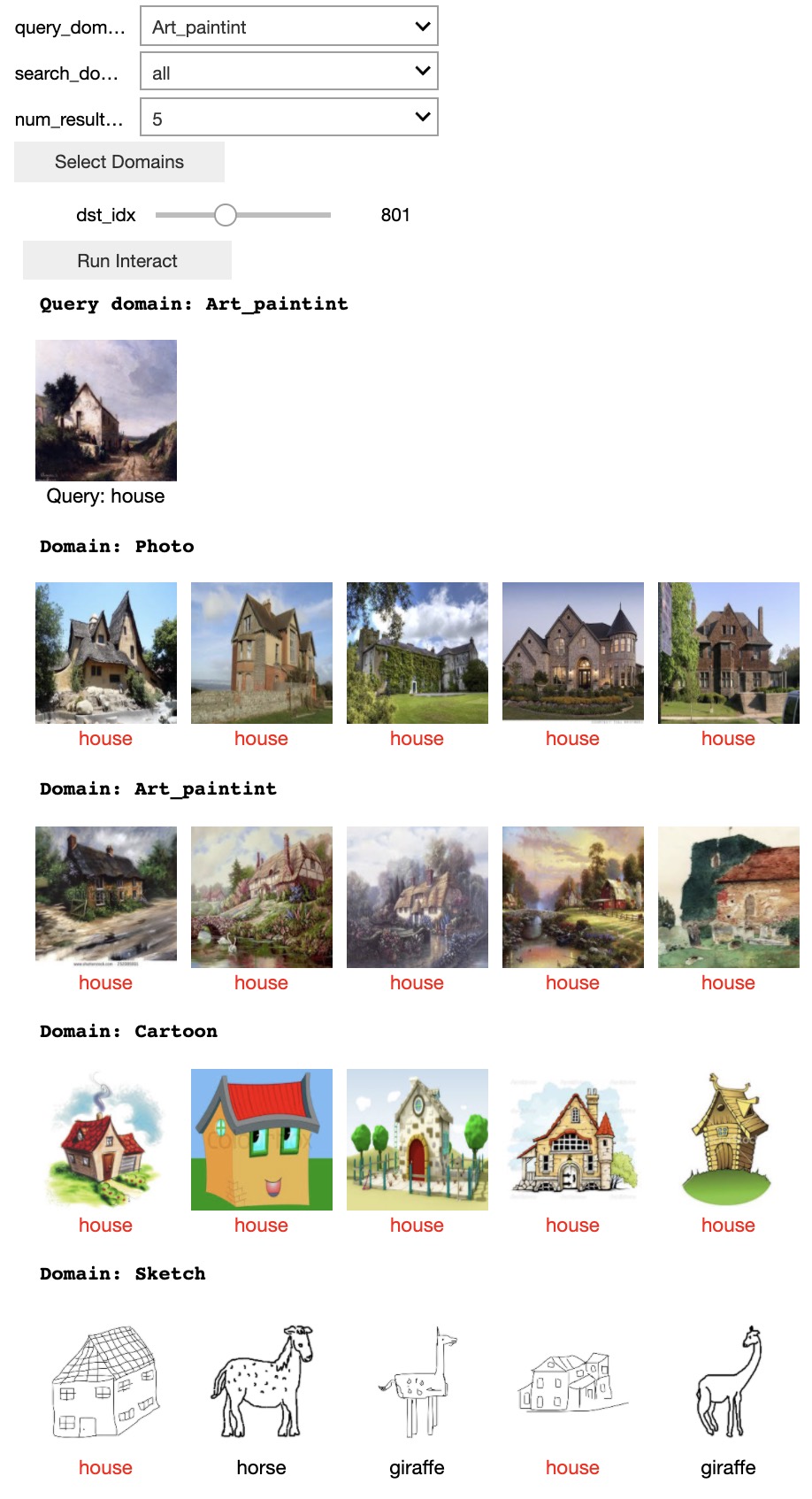}
    \caption{\democap{House}}
    \label{fig:demo13}
\end{figure}

\begin{figure}
    \centering
    \includegraphics[width=\linewidth]{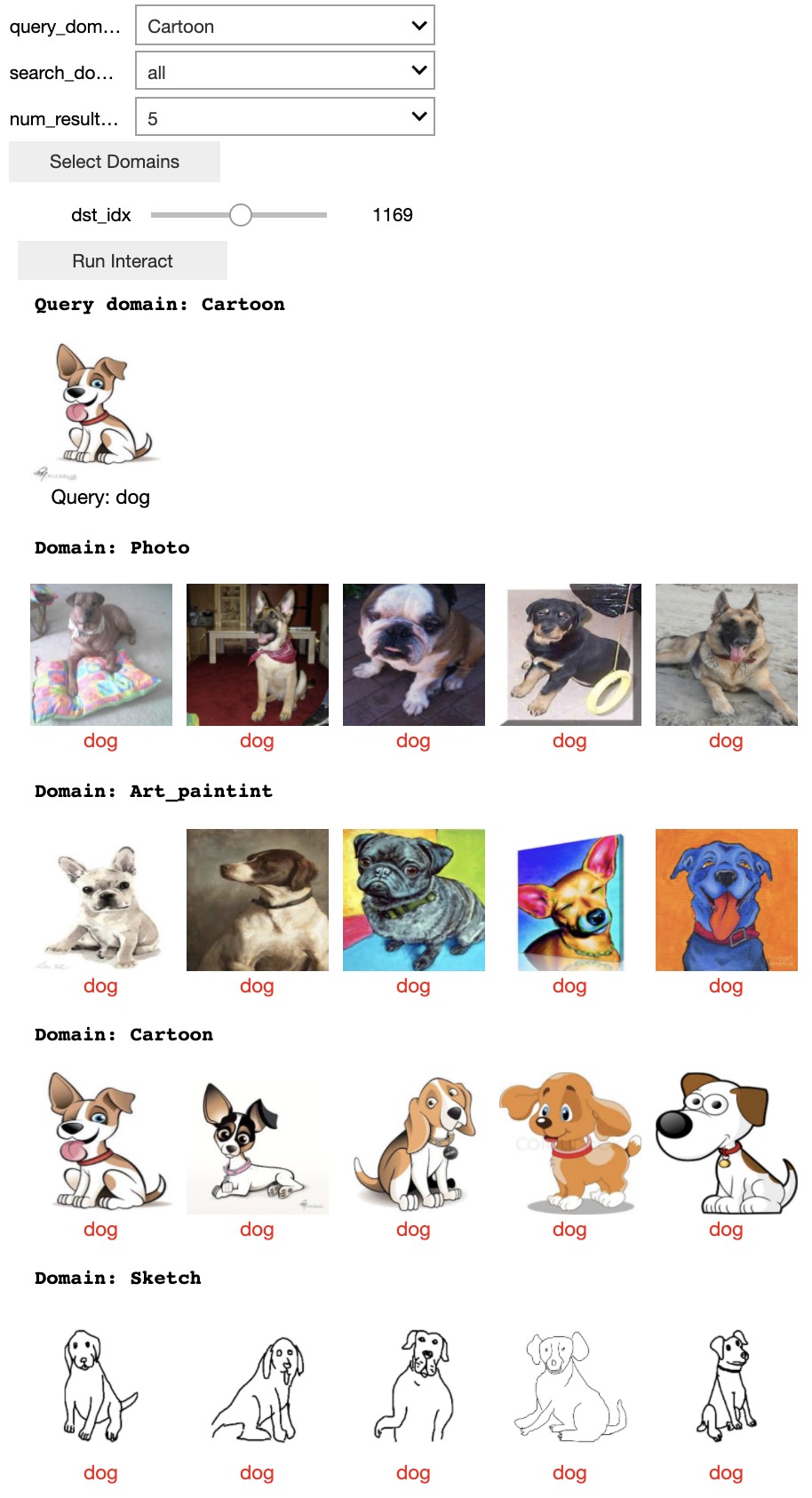}
    \caption{\democap{Dog}}
    \label{fig:demo14}
\end{figure}

\begin{figure}
    \centering
    \includegraphics[width=\linewidth]{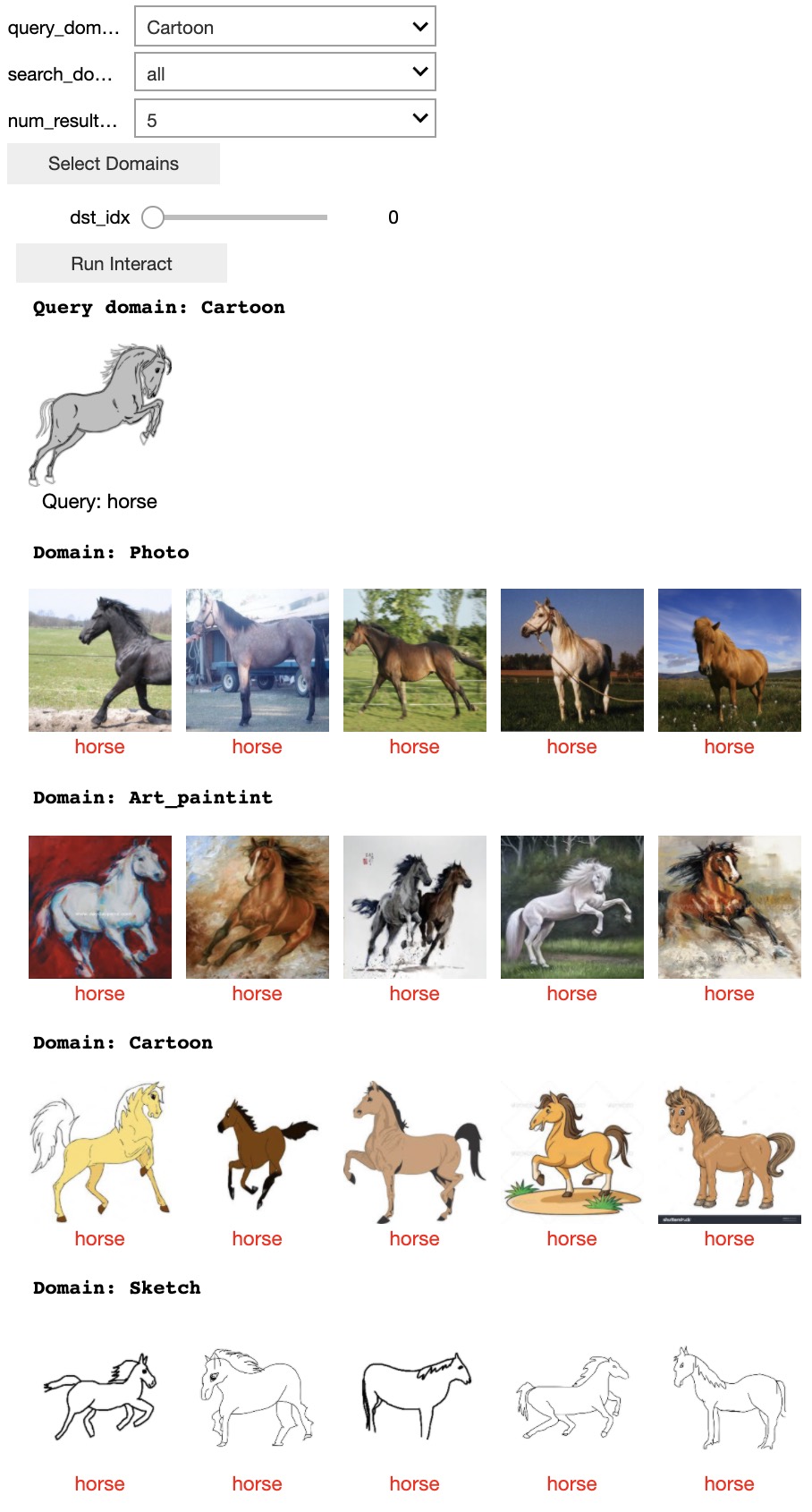}
    \caption{\democap{Horse}}
    \label{fig:demo15}
\end{figure}

\begin{figure}
    \centering
    \includegraphics[width=\linewidth]{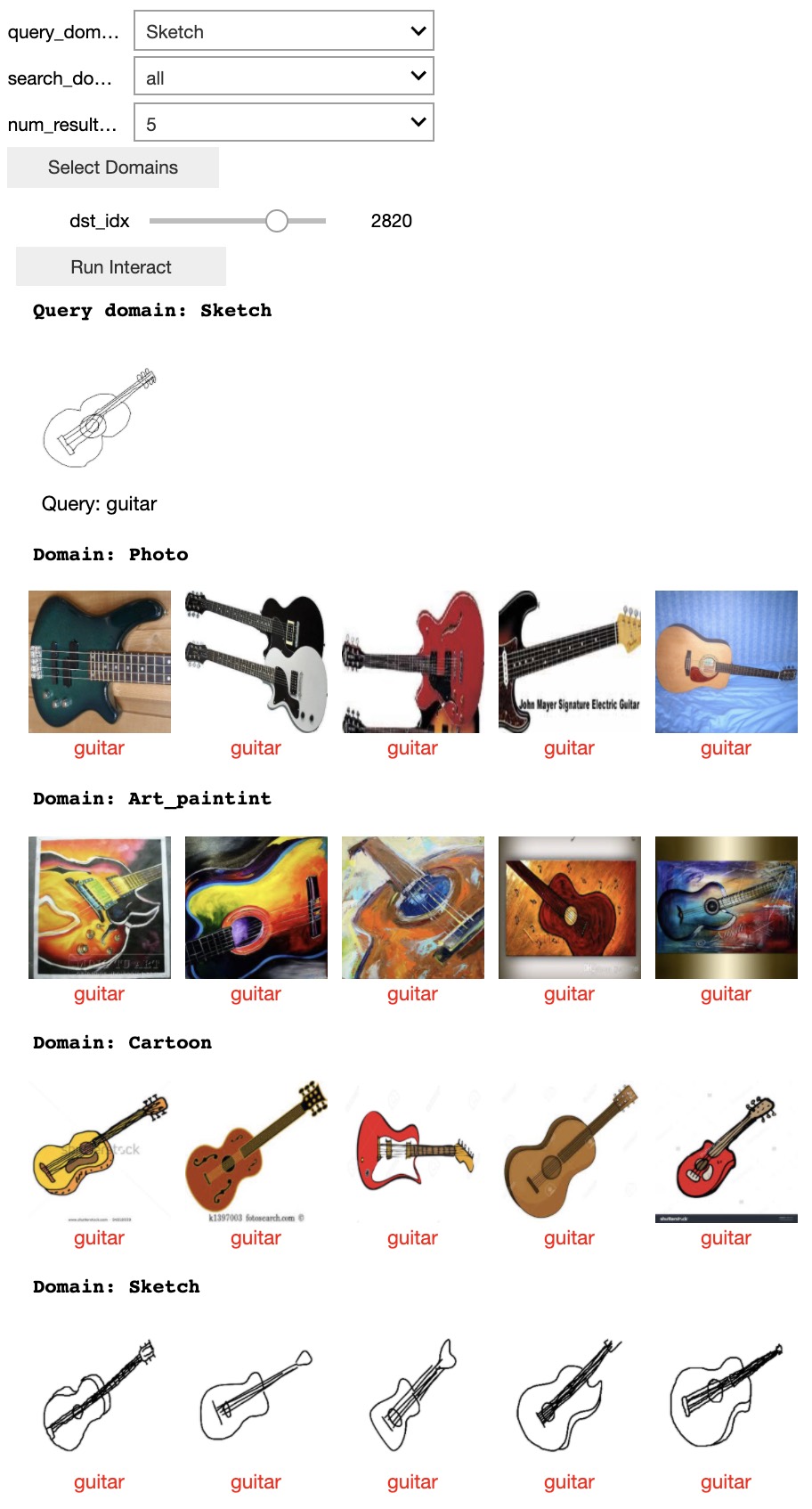}
    \caption{\democap{Guitar}}
    \label{fig:demo16}
\end{figure}

In in \cref{fig:demo1,fig:demo2,fig:demo3,fig:demo4,fig:demo5,fig:demo6,fig:demo7,fig:demo8,fig:demo9,fig:demo10,fig:demo11,fig:demo12,fig:demo13,fig:demo14,fig:demo15,fig:demo16} we showcase the domain alignment capabilities of the feature representation learned without supervision using our \oursspace approach. Each example shows top-$5$ nearest neighbors of a random query image (from the PACS dataset) searched in the entire set of images of each of the $4$ different PACS domains: Photo, Art/Painting, Cartoon, and Sketch. All images are encoded using our self-supervised \oursspace model trained on DomainNet dataset.

\end{document}